\documentclass[letterpaper]{article} 
\usepackage{aaai2026}  
\usepackage{times}  
\usepackage{helvet}  
\usepackage{courier}  
\usepackage[hyphens]{url}  
\usepackage{graphicx} 
\usepackage{natbib}  
\usepackage{caption} 

\usepackage{algorithm}
\usepackage{algorithmic}

\usepackage{newfloat}
\usepackage{listings}
\DeclareCaptionStyle{ruled}{labelfont=normalfont,labelsep=colon,strut=off} 
\floatstyle{ruled}
\newfloat{listing}{tb}{lst}{}
\floatname{listing}{Listing}
\usepackage[utf8]{inputenc} 
\usepackage[T1]{fontenc}    
\usepackage{url}            
\usepackage{booktabs}       
\usepackage{amsfonts}       
\usepackage{nicefrac}       
\usepackage{microtype}      
\usepackage[cmyk,table]{xcolor}        
\usepackage{array}
\usepackage{cite}
\PassOptionsToPackage{numbers, compress}{natbib}
\usepackage{enumitem}
\usepackage{threeparttable}
\usepackage{comment}
\usepackage{microtype}
\usepackage{multicol}
\usepackage{multirow}
\usepackage{amsmath,amsthm,amsfonts,amssymb}
\usepackage{mathtools}
\usepackage{tkz-euclide} 
\usepackage{siunitx}
\usepackage{tikz}
\usetikzlibrary{fit,calc,positioning,backgrounds,shapes.geometric,decorations.pathreplacing}
\usepackage{bm}
\usepackage{tikz-3dplot}
\usepackage{subcaption,color}
\usetikzlibrary{arrows.meta}
\usepackage{arydshln}
\allowdisplaybreaks

\usepackage{threeparttable} 

\usepackage{makecell}
\theoremstyle{plain}
\newtheorem{theorem}{Theorem}[section]

\theoremstyle{definition}

\theoremstyle{remark}

\newcommand{\nn}{\num[group-separator={,},group-minimum-digits=3]}
\newcommand{\B}{\bfseries}

\usepackage[disable,textsize=tiny]{todonotes}

\definecolor{NavyBlue}{cmyk}{0.65,0.25,0,0.05}
\definecolor{TableBlue}{cmyk}{0.35,0.05,0,0.02} 
\definecolor{RegionYellow}{cmyk}{0, 0.05, 0.25, 0}
\definecolor{StepBoxPink}{cmyk}{0, 0.6, 0.8, 0.3}
\definecolor{StepBoxPinkLight}{cmyk}{0, 0.2, 0.3, 0.05}
\definecolor{LightGrey}{cmyk}{0, 0, 0, 0.5}
\tikzset{
	reuse path/.code={\pgfsyssoftpath@setcurrentpath{#1}}
}
\tikzset{even odd clip/.code={\pgfseteorule},
	protect/.code={
		\clip[overlay,even odd clip,reuse path=#1]
		(-6383.99999pt,-6383.99999pt) rectangle (6383.99999pt,6383.99999pt);
}}
\tikzset{
	dot/.style={circle, fill, minimum size=#1, inner sep=0pt, outer sep=0pt},
	dot/.default = 4.5pt,
	hemispherebehind/.style={ball color=gray!20!white, fill=none, opacity=0.3},
	hemispherefront/.style={ball color=gray!65!white, fill=none, opacity=0.4},
	ellipsoidfront/.style={ball color=gray!50!white, fill=none, opacity=0.5},
	circlearc/.style={thick,color=gray!90},
	circlearchidden/.style={thick,dashed,color=gray!90},
	equator/.style = {thick, black},
	diameter/.style = {thick, black},
	axis/.style={thick, -stealth,black!60, every node/.style={text=black, at={([turn]1mm,0mm)}},
	},
}

\newcommand{\bx}{\mathbf{x}}
\newcommand{\bu}{\mathbf{u}}

\newcommand{\bv}{\mathbf{v}}

\newcommand{\bp}{\mathbf{p}}

\newcommand{\bg}{\mathbf{g}}
\newcommand{\E}{\mathbb{E}}

\title{Improving the Convergence Rate of Ray Search Optimization for\\Query-Efficient Hard-Label Attacks}
\author{
    Xinjie Xu\textsuperscript{\rm 1},
    Shuyu Cheng\textsuperscript{\rm 2},
    Dongwei Xu\textsuperscript{\rm 1},
    Qi Xuan\textsuperscript{\rm 1,3},
    Chen Ma\textsuperscript{\rm 1,3}\thanks{Corresponding author.}
}
\affiliations{
    \textsuperscript{\rm 1}Institute of Cyberspace Security, Zhejiang University of Technology, Hangzhou 310023, China\\
	\textsuperscript{\rm 2}JQ Investments, Shanghai 200122, China\\
	\textsuperscript{\rm 3}Binjiang Institute of Artificial Intelligence, ZJUT, Hangzhou 310056, China\\
    xxj1018@foxmail.com, csy530216@126.com, \{dongweixu, xuanqi, machen\}@zjut.edu.cn
}

\begin{document}

\maketitle

\begin{abstract}
In hard-label black-box adversarial attacks, where only the top-1 predicted label is accessible, the prohibitive query complexity poses a major obstacle to practical deployment. In this paper, we focus on optimizing a representative class of attacks that search for the optimal ray direction yielding the minimum $\ell_2$-norm perturbation required to move a benign image into the adversarial region. Inspired by Nesterov's Accelerated Gradient (NAG), we propose a momentum-based algorithm, ARS-OPT, which proactively estimates the gradient with respect to a future ray direction inferred from accumulated momentum. We provide a theoretical analysis of its convergence behavior, showing that ARS-OPT enables more accurate directional updates and achieves faster, more stable optimization. To further accelerate convergence, we incorporate surrogate-model priors into ARS-OPT's gradient estimation, resulting in PARS-OPT with enhanced performance. The superiority of our approach is supported by theoretical guarantees under standard assumptions. Extensive experiments on ImageNet and CIFAR-10 demonstrate that our method surpasses 13 state-of-the-art approaches in query efficiency.
\end{abstract}
 \begin{links}
     \link{Code}{https://github.com/machanic/hard_label_attacks}
 \end{links}

\section{Introduction}
\label{sec:intro}

	We focus on hard-label adversarial attacks. Considered among the most practical and challenging black-box attacks, hard-label attacks operate under strict information constraints. While white-box attacks \cite{goodfellow6572explaining, madry2018towards} leverage model parameters and gradients, and score-based attacks \cite{ma2021simulator} exploit confidence scores, hard-label attacks rely solely on top-1 predicted labels. This makes the efficient generation of adversarial examples substantially more difficult while enhancing their practical applicability.
	
	\textbf{Why study query-based black-box adversarial attacks under the hard-label setting?} Real-world machine-learning services such as cloud vision APIs and biometric recognizers often reveal nothing more than the final predicted decision (i.e., the top-1 label) to external users. 
	With gradients and confidence scores stripped away, an attacker is forced to treat the model as a hard‑label black box to probe its decision boundary. This stringent setting accurately reflects the limited feedback of deployed services and raises three key challenges. (1) \emph{Minimal feedback:} Each query yields only a hard-label response, demanding efficient exploration strategies. (2) \emph{Practical relevance:} It closely mirrors restricted commercial platforms where probability scores and internal details are deliberately hidden. (3) \emph{Security-critical:} Hard-label attacks reveal vulnerabilities in ``security-through-obscurity'' systems and underscore the urgent need for defenses against adversaries with minimal information. Consequently, designing query-efficient attacks based solely on hard-label feedback is essential for vulnerability assessment and robust defenses.
	
	\textbf{Why are hard-label attacks challenging?} Because a model's predicted label typically changes only when an input moves across or near its decision boundary, hard-label attacks must restrict their search to this narrow region, making the optimization especially challenging.
	Early hard-label attacks like Boundary Attack (BA) \cite{brendel2018decisionbased} and Biased BA \cite{brunner2019guessing} initialize from a sample already in the adversarial region and progressively reduce the perturbation by stepping toward the original image while exploring directions on the decision boundary via randomly sampled spherical vectors. However, these approaches remain highly inefficient in terms of query cost: they rely almost entirely on random sampling and neglect valuable information from past queries, which impedes effective perturbation reduction. To address this challenge, recent studies have adopted zeroth-order (ZO) optimization techniques, which leverage boundary information more effectively to identify adversarial examples. Existing ZO-based attacks---such as HopSkipJumpAttack (HSJA) \cite{chen2019hopskipjumpattack}, OPT \cite{cheng2019queryefficient}, Sign-OPT \cite{cheng2020sign}, and Prior-OPT \cite{ma2025boosting}---primarily focus on improving gradient estimation through finite differences. However, their optimization strategies rely on vanilla gradient descent, overlooking well-established acceleration methods such as momentum and Nesterov's accelerated gradient, which can enhance convergence rates even when the gradient estimation quality remains unchanged. To address these limitations, we propose ARS-OPT, a novel ZO optimization algorithm incorporating accelerated random search (ARS) \cite{nesterov2017random}. Our theoretical analysis demonstrates that ARS-OPT leverages second-order gradient information implicitly without requiring explicit Hessian estimation and establishes a bound on the expected gap between the objective value at iteration $T$ and the optimum value. Building on this, we introduce PARS-OPT, which integrates transfer-based priors to improve gradient estimation. PARS-OPT further extends to combine priors from multiple surrogate models, delivering additional gains in attack performance. Extensive experiments on ImageNet, CIFAR-10, and a CLIP-based model demonstrate that our framework, consisting of ARS-OPT and its prior-enhanced variant PARS-OPT, outperforms 13 state-of-the-art baseline methods with superior query efficiency.
	
    Our main contributions are summarized as follows.
    \begin{itemize}
        \item \textbf{Novelty in hard-label attacks.} We present ARS-OPT, a novel hard-label attack that accelerates convergence by estimating gradients along an interpolated ``lookahead'' direction, combining the search trajectory with accumulated momentum. We further introduce PARS-OPT, which integrates transfer-based priors from surrogate models to improve gradient estimation and enhance attack efficiency.
        \item \textbf{Novelty in theoretical analysis.} We establish an $\mathcal{O}(1/T^2)$ convergence rate under standard assumptions, supported by the construction of an unbiased estimator of the true gradient that is essential for ensuring this rate. The theoretical analysis provides a principled explanation for the acceleration behavior of our approach and clarifies its underlying optimization dynamics.
        \item \textbf{SOTA performance.} Experimental results show our approach outperforms 13 state-of-the-art attacks on ImageNet and CIFAR-10 across classifiers, including CLIP.
    \end{itemize}

\section{Related Work}
Hard-label attacks, also known as decision-based black-box attacks, are among the most challenging adversarial scenarios. They rely solely on the target model's top-1 predicted label without access to internal structure or confidence scores, and craft perturbations by querying and exploiting information near the decision boundary.
Boundary Attack (BA) \cite{brendel2018decisionbased} was one of the earliest methods, performing random walks on the boundary to minimize perturbations, but suffers from low query efficiency. Biased BA (BBA) \cite{brunner2019guessing} improves BA via three biases: (1) low-frequency Perlin noise, (2) regional masking, and (3) surrogate-model gradients. 
The Evolutionary Attack (abbreviated as Evolutionary) \cite{dong2019efficient} adopts random sampling with adaptive covariance, while AHA \cite{li2021aha} exploits historical queries to guide the search.
HopSkipJumpAttack (HSJA) \cite{chen2019hopskipjumpattack} refines adversarial examples via (1) gradient approximation at the boundary and (2) binary search projection onto the boundary toward the benign image. 
SQBA \cite{park2024sqba} combines surrogate-model gradients with HSJA's gradient estimation to improve query efficiency.
QEBA \cite{li2020qeba} lowers HSJA's query cost using subspaces derived from spatial transformations, low-frequency components, and intrinsic features. GeoDA \cite{rahmati2020geoda} leverages the boundary's low curvature via local linearization to estimate gradients and reduce queries. Triangle Attack \cite{wang2022triangle} applies the law of sines in a low-frequency subspace, removing boundary projections and gradient estimation. Tangent Attack (TA) \cite{ma2021finding} locates an optimal tangent point to minimize perturbations, while SurFree \cite{maho2021surfree} uses geometry-driven directional trials without gradient estimation. CGBA and its variant CGBA-H \cite{reza2023cgba} search along a semicircular path on a restricted 2D plane to find boundary points.
Another direction formulates hard-label attacks as continuous optimization problems. OPT \cite{cheng2019queryefficient} employs zeroth-order (ZO) optimization based on random-direction finite differences. Sign-OPT \cite{cheng2020sign} reduces queries by using directional derivative signs but sacrifices gradient precision. Prior-OPT \cite{ma2025boosting} integrates transfer-based priors into the ray-search optimization, while RayS \cite{chen2020rays} removes gradient estimation entirely by using hierarchical search, but is limited to untargeted $\ell_\infty$-norm attacks. QE-DBA \cite{zhang2024qedba} applies Bayesian optimization to explore the perturbation space, effectively addressing hard-label ZO optimization problems.
However, existing methods overlook established acceleration strategies---such as momentum and Nesterov's accelerated gradient---that can greatly improve convergence rates without requiring better gradient estimates. In this work, we address this gap by integrating acceleration techniques to enhance query efficiency. Moreover, our framework can further boost efficiency by incorporating transfer-based priors.

\section{Problem Statement of Hard-Label Attacks}
Given a classifier $\psi \vcentcolon \mathbb{R}^d \rightarrow \mathbb{R}^{C}$ designed for a $C$-class classification task, and a correctly classified input image $\bx \in [0,1]^d$, where $d$ is the dimension of the input image, the adversary seeks to generate an adversarial example $\bx_\text{adv}$ by crafting a minimal perturbation such that the classifier's prediction for $\bx_\text{adv}$ becomes incorrect.
This adversarial objective can be formally expressed as:
\begin{small}
\begin{equation}
    \label{eq:goal_hard_label}
    \min_{\mathbf{x}_\text{adv}} \,\| \bx_\text{adv} - \bx \|_p \quad \text{s.t. }\quad \Phi(\mathbf{x}_\text{adv}) = 1,
\end{equation}
\end{small}
where $\| \bx_\text{adv} - \bx \|_p$ is the $p$-norm distortion, and the constraint $\Phi(\mathbf{x}_\text{adv})$ is defined as an attack success indicator:
\begin{footnotesize}
\begin{equation}
    \Phi(\mathbf{x}_{\text{adv}}) \coloneqq \begin{cases}
    1 & \text{if } \hat{y} = y_\text{adv}\text{ in a targeted attack},\\
    & \quad\text{or } \hat{y} \neq y \text{ in an untargeted attack},\\
    0 & \text{otherwise}.
    \end{cases}
\label{eq:phi}
\end{equation}
\end{footnotesize}
Here, $\hat{y} = \arg \max_{i\in \{1,\dots,C\}} \psi(\mathbf{x}_{\text{adv}})_i$ denotes the top-1 predicted label by classifier $\psi$, $y$ is the true label of $\bx$, and $y_\text{adv}$ is the target label in a targeted attack scenario.

Following the ray-search methods \cite{cheng2019queryefficient,cheng2020sign,ma2025boosting}, we reformulate the optimization problem in Eq. \eqref{eq:goal_hard_label} as finding the optimal ray direction $\theta^*$ from $\bx$ that yields the minimal distance $f(\theta)$ to the boundary of the adversarial region. This can be formulated as:
\begin{footnotesize}
\begin{equation}
\begin{gathered}
    \label{eq:goal_OPT}
    \min_{\theta \in \mathbb{R}^d\setminus\{\mathbf{0}\}} f(\theta), \\
    \text{where}\quad f(\theta) \coloneqq \inf\Bigl\{ \lambda>0: \Phi\bigl(\mathbf{x} + \lambda \frac{\theta}{\|\theta\|}\bigr) = 1 \Bigr\}.
\end{gathered}
\end{equation}
\end{footnotesize}

By convention, $f(\theta) = +\infty$ if the set is empty.
Consequently, the resulting adversarial example is constructed as $\bx^* = \bx + f(\theta^*) \frac{\theta^*}{\|\theta^*\|}$, where $\theta^*$ is the optimal solution obtained from the minimization problem defined in Eq. \eqref{eq:goal_OPT}.

\section{The Proposed Approach}
Previous works \cite{cheng2019queryefficient, cheng2020sign, ma2025boosting} focus on efficient gradient estimation to optimize the direction $\theta$, with step size typically determined by line search. However, they do not explore any optimization acceleration techniques beyond gradient estimation. Next, we present an overview of ARS-OPT and its prior-enhanced variant PARS-OPT, both equipped with theoretical convergence guarantees.
	
\subsection{Conceptual Sketch and Overview}
\label{sec:sketch}
\citet{nesterov2017random} propose an Accelerated Random Search (ARS) method for ZO optimization, which rigorously establishes explicit non-asymptotic convergence rates under various convexity and smoothness assumptions by introducing an accelerated ZO framework. In \emph{the score-based setting}, Cheng et al. \citeyearpar{cheng2021ontheconvergence} extend ARS to score-based attacks and provide an analysis of the convergence rate.
However, in \emph{hard-label attacks}, obtaining function values requires extensive binary searches, significantly reducing the query efficiency of gradient estimation based on finite differences.

\begin{figure}
	\centering
	\def\r{1}
	\def\delta{5}   
	\begin{tikzpicture}[scale=1.5,myarrow/.style={-{Stealth[length=0.8mm, width=0.8mm]}, line width=0.5pt},tinyarrow/.style={{Stealth[length=0.5mm, width=0.6mm]}-, line width=0.5pt},tinyarrow2/.style={-{Stealth[length=0.5mm, width=0.6mm]}, line width=0.5pt}, inner sep=0pt, outer sep=0pt]	
		\coordinate (start_curve) at (2.3,4.6);
		\coordinate (lowest_point) at (3.8,3.4);
		\coordinate (end_curve) at (4.7,4.6);
		\path[name path=adv_boundary,rounded corners] (start_curve) -- (lowest_point) --  (end_curve);

	\shade[
	shading = radial,
	inner color = white,      
	outer color = NavyBlue,    
	shading angle=0,
	opacity = 0.9             
	]
	(start_curve) parabola bend (lowest_point) (end_curve);

	\node at ($(start_curve)!0.5!(end_curve) + (0.2,-0.3)$)[font=\small] {adversarial region};
	\draw[very thick] (start_curve) parabola bend (lowest_point) (end_curve);
	\path[name path=curve](start_curve) parabola bend (lowest_point) (end_curve);
	\pgfresetboundingbox \path (start_curve); \path (end_curve); 
		\coordinate (O) at  (3.2,2.1);

		\path[name path=mycircle]
		(O) ++(180+\delta:\r)    
		arc[
		start angle=180+\delta,
		end angle=-10,  
		radius=\r
		];
		
		\coordinate (theta_t) at (2.7,4);
		\coordinate (tilde_theta_t) at (3,4);
		\coordinate (theta_t_plus_1) at (3.5,4);
		\coordinate (m_t) at (4.7,4);
		\coordinate (m_t_plus_1) at (5.1,4);

		\path[name path=ray_theta_t] (O) -- ($(O)!1.1!(theta_t)$);
		\path[name intersections={of=ray_theta_t and curve, by=I_theta_t}];
		\path[name intersections={of=mycircle and ray_theta_t, by=IC_theta_t}];

		\path[name path=ray_tilde_theta_t] (O) -- (tilde_theta_t);
		\path[name intersections={of=ray_tilde_theta_t and curve, by=I_tilde_theta_t}];
		
		\coordinate (g1) at ($ (I_tilde_theta_t) !{\r cm}! ($(I_tilde_theta_t)+(-0.4,-0.2)$) $);
		\draw[-stealth] (I_tilde_theta_t) -- (g1) node[anchor=south east,font=\scriptsize]{$g_1(\tilde{\theta}_t)$};

		\coordinate (g2) at ($ (I_tilde_theta_t) !{1.22\r cm}! ($(I_tilde_theta_t) + (-0.7, -0.6)$) $);
		\draw[-stealth] (I_tilde_theta_t) -- (g2) node[anchor=east,font=\scriptsize]{$g_2(\tilde{\theta}_t)$};
		
		\draw ($(g2) + (0.1,-0.3)$) node[font=\scriptsize,fill=gray!20,
		inner sep=2pt,
		rounded corners=2pt] {$\mathbb{E}[g_2(\tilde{\theta}_t)]=\nabla f(\tilde{\theta}_t)$};
		
		\draw ($(O) + (1.4\r,0)$) node[font=\small]{non-adversarial region};
		\path[name intersections={of=mycircle and ray_tilde_theta_t, by=IC_tilde_theta_t}];

		\path[name path=ray_theta_t_plus_1] (O) -- (theta_t_plus_1);
		\path[name intersections={of=ray_theta_t_plus_1 and curve, by=I_theta_t_plus_1}];
		\path[name intersections={of=mycircle and ray_theta_t_plus_1, by=IC_theta_t_plus_1}];
		\pic [draw=NavyBlue,text=black!50!green, angle eccentricity=1.4,angle radius=0.8cm,tinyarrow,line width=1pt,{Stealth[length=0.8mm, width=0.9mm]}-] {angle = IC_theta_t_plus_1--O--IC_tilde_theta_t};
		
		\draw[densely dashed,NavyBlue] (IC_theta_t_plus_1) -- (I_theta_t_plus_1)node[anchor=south west, font=\scriptsize,color=black,xshift=2pt]{$f(\theta_{t+1})$};

		\path[name path=ray_m_t] (O) -- (m_t);
		\path[name intersections={of=mycircle and ray_m_t, by=IC_m_t}];
		
		\path[name path=ray_m_t_plus_1] (O) -- (m_t_plus_1);
		\path[name intersections={of=mycircle and ray_m_t_plus_1, by=IC_m_t_plus_1}];
		
		\draw[-stealth, NavyBlue] (O) -- (IC_theta_t) node[anchor=north east,yshift=-2pt,font=\scriptsize]{$\theta_t$};
		\draw[-stealth, NavyBlue] (O) -- (IC_tilde_theta_t) node[anchor=north west,xshift=1pt,yshift=-1pt,font=\scriptsize]{$\tilde{\theta}_t$};

		\draw[-stealth, NavyBlue] (O) -- (IC_theta_t_plus_1) node[anchor=north west,xshift=1pt,yshift=-2pt,font=\scriptsize]{$\theta_{t+1}$};
		\draw[-stealth, black!50!green] (O) -- (IC_m_t) node[anchor=north east,xshift=-3pt,yshift=-1pt,font=\scriptsize]{$m_t$};
		\draw[-stealth, black!50!green] (O) -- (IC_m_t_plus_1)  node[anchor=north west,xshift=1pt,yshift=-1pt,font=\scriptsize]{$m_{t+1}$};
		\pic [draw=black!50!green,text=black!50!green, angle eccentricity=1.5,angle radius=1cm,tinyarrow,line width=1pt,{Stealth[length=0.8mm, width=0.9mm]}-] {angle = IC_m_t_plus_1--O--IC_m_t};

		\draw[densely dashed,NavyBlue] (IC_theta_t) -- (I_theta_t) node[anchor=south west, font=\scriptsize,color=black,xshift=2pt]{$f(\theta_t)$};
		\fill[orange] (I_theta_t) circle (1pt);
		\draw[densely dashed,NavyBlue] (IC_tilde_theta_t) -- (I_tilde_theta_t) node[anchor=south west, font=\scriptsize,color=black,xshift=2pt]{$f(\tilde{\theta}_t)$};
		\fill[orange] (I_tilde_theta_t) circle (1pt);
		
		\draw[densely dashed,NavyBlue] (IC_theta_t_plus_1) -- (I_theta_t_plus_1)node[anchor=south west, font=\scriptsize,color=black,xshift=2pt]{$f(\theta_{t+1})$};
		\fill[orange] (I_theta_t_plus_1) circle (1pt);
		\coordinate (m_t_plus_1_raw) at ($(O)!{3*\r cm}!(m_t_plus_1)$);

		\coordinate (to_new_theta) at ($ (IC_tilde_theta_t) + (I_tilde_theta_t) - (g1) $);
		\path[name path=ray_g1] (IC_tilde_theta_t) -- (to_new_theta);
		\path[name path=ray_g1_end] (IC_theta_t_plus_1) -- (I_theta_t_plus_1);
		\path[name intersections={of=ray_g1 and ray_g1_end, by=theta_t_plus_1_raw}];
		\draw[-stealth] (IC_tilde_theta_t) -- (theta_t_plus_1_raw) node[anchor=south,font=\scriptsize, yshift=-2.5pt,xshift=-4pt]{$-\frac{1}{\hat{L}}g_1(\tilde{\theta}_t)$};
		
		\coordinate (to_new_m) at ($ (IC_m_t) + 2*(I_tilde_theta_t) - 2*(g2) $);
		\path[name path=ray_g2] (IC_m_t) -- (to_new_m);
		\path[name path=ray_g2_end] (IC_m_t_plus_1) -- (m_t_plus_1_raw);
		\path[name intersections={of=ray_g2 and ray_g2_end, by=new_i}];
		\draw[-stealth] (IC_m_t) -- (new_i) node[anchor=south,font=\scriptsize, yshift=-2.5pt,xshift=-4pt]{$-\frac{\zeta_t}{\alpha_t}g_2(\tilde{\theta}_t)$};
		\draw[densely dashed] (IC_m_t_plus_1) -- (new_i);

		\draw ($(g1) + (-0.6,0.6)$) node[font=\scriptsize,fill=gray!20,
	inner sep=1pt, outer sep=0pt,align=left,
	rounded corners=2pt] {In ARS-OPT, $g_1(\tilde{\theta}_t)$ and \\
		$g_2(\tilde{\theta}_t)$ are collinear, but this\\does not hold in PARS-OPT.};
		
		\draw[-stealth] ($(O) + (\r,0.4)$) -- ++(0.2, 0) 
		node[right, font=\scriptsize, align=left,fill=gray!20,
		inner sep=2pt, outer sep=0pt,align=left,
		rounded corners=2pt] {
			The circle represents\\the unit-norm constraint.
		};

		\draw[very thick]
	(current bounding box.south west)
	rectangle
	(current bounding box.north east);
	\coordinate (bb-sw) at (current bounding box.south west);
	\coordinate (bb-ne) at (current bounding box.north east);

	\fill[fill=black] (O) circle[radius=1pt] node[anchor=east,xshift=-2pt,font=\small]{original image};
	\begin{scope}
		\clip (bb-sw) rectangle (bb-ne);
		\draw[densely dashed] (O) circle (\r);
	\end{scope}
	
	\end{tikzpicture}
	\caption{Illustration of a three-step update: first, compute the perturbation direction $\tilde{\theta}_t = (1-\alpha_t)\theta_t + \alpha_t m_t$; then estimate gradients at $\tilde{\theta}_t$ using a biased $g_1(\tilde{\theta}_t)$ and an unbiased $g_2(\tilde{\theta}_t)$; finally, update $\theta_{t+1}$ and $m_{t+1}$ via a gradient descent step.}
	\label{fig:three-update-step}
\end{figure}
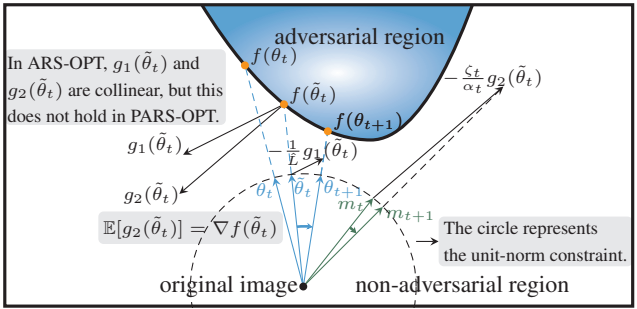
To address these limitations, we introduce ARS-OPT, a novel ZO optimization framework that can be seamlessly augmented with transfer-based priors to further boost query efficiency.
The primary challenge is accelerating convergence in gradient descent when only poorly estimated gradients are available. At iteration $t$, we employ the following three‐step update process for $\theta_t$ (Fig. \ref{fig:three-update-step}):

\begin{enumerate}[label=\arabic*,ref=\arabic*]
    \item \label{step:perturb} Compute the perturbation direction $\tilde{\theta}_t \gets (1 - \alpha_t)\theta_t + \alpha_t m_t$, where $m_0$ is initialized to $\theta_0$.
    \item \label{step:estimate} At $\tilde{\theta}_t$, we use multiple queries to estimate gradients $g_1(\tilde{\theta}_t)$ (biased estimator, e.g., Sign-OPT or Prior-OPT method) and $g_2(\tilde{\theta}_t)$ (unbiased estimator of $\nabla f(\tilde{\theta}_t)$).
    \item \label{step:update} Update both parameters by gradient descent: $\theta_{t+1} \gets \tilde{\theta}_t - \frac{1}{\hat{L}} g_1(\tilde{\theta}_t)$, $m_{t+1} \gets m_t - \frac{\zeta_t}{\alpha_t}g_2(\tilde{\theta}_t)$.
\end{enumerate}

Inspired by Nesterov's accelerated gradient method, our approach dynamically tracks two sequences, i.e., the direction $\theta_t$ and the momentum vector $m_t$, and then computes a lookahead vector $\tilde{\theta}_t$ by linearly interpolating between $\theta_t$ and $m_t$, controlled by an interpolation coefficient $\alpha_t$. At $\tilde{\theta}_t$, we estimate two gradients, $g_1(\tilde{\theta}_t)$ and $g_2(\tilde{\theta}_t)$, to compute the updates of $\theta_{t+1}$ and $m_{t+1}$, respectively. Although we adopt the same estimation procedure for $g_1(\tilde{\theta}_t)$ as in Prior-OPT, our algorithm converges substantially faster, as demonstrated by our experiments. The convergence guarantee of our approach relies on two technical assumptions: (1) $g_{2}(\tilde{\theta}_{t})$ serves as an unbiased estimator of $\nabla f(\tilde{\theta}_{t})$, and (2) $\zeta_{t} \le \mathbb{E}_{t}\!\left[(\nabla f(\tilde{\theta}_{t})^{\top} \bv_{t})^{2}\right] / \left(\hat{L} \cdot \mathbb{E}_{t}\!\left[\|g_{2}(\tilde{\theta}_{t})\|^2\right]\right)$, with the full derivation given in the Appendix. We also note that our framework can incorporate various gradient estimation techniques, such as prior-guided estimation, to further improve performance. Our approach can be intuitively understood through the analogy of a walker descending a valley: rather than relying solely on the current slope, the walker looks ahead to anticipate the upcoming terrain and adjust the direction of motion accordingly, thereby achieving smoother and faster progress toward the minimum.

\begin{figure*}[t]
	\centering
	\begin{tikzpicture}[
		font=\normalsize,
		box/.style={draw, thick, minimum height=1.2cm, minimum width=1.8cm, align=center},
		imgbox/.style={draw, very thick, align=center, rounded corners=4pt,inner sep=1pt},
		imgbox2/.style={draw, very thick, align=center, rounded corners=4pt,inner xsep=1.5pt,inner ysep=1pt},
		imgbox_horizontal_narrow/.style={draw, very thick, align=center, rounded corners=4pt,inner xsep=1pt,inner ysep=1pt},
		imgbox_all_narrow/.style={draw, very thick, align=center, rounded corners=4pt,inner xsep=0pt,inner ysep=1pt},
		arrow/.style={thick, -{Latex[length=2.5mm]}},
		]
	
\node[imgbox] (img) {
	
	\begin{tikzpicture}
		\matrix[
		ampersand replacement=\&,
		column sep=2pt,
		row sep=0pt, nodes={anchor=center}  
		](m) {
			\node[
			align=center,
			inner sep=0pt,
			minimum width=1.65cm
			] (orig_label) {Original\\Image\\$\mathbf{x}$}; 
			\&
			\node[
			align=center,
			inner sep=1pt
			] (orig_img) {
				\includegraphics[width=1.5cm]{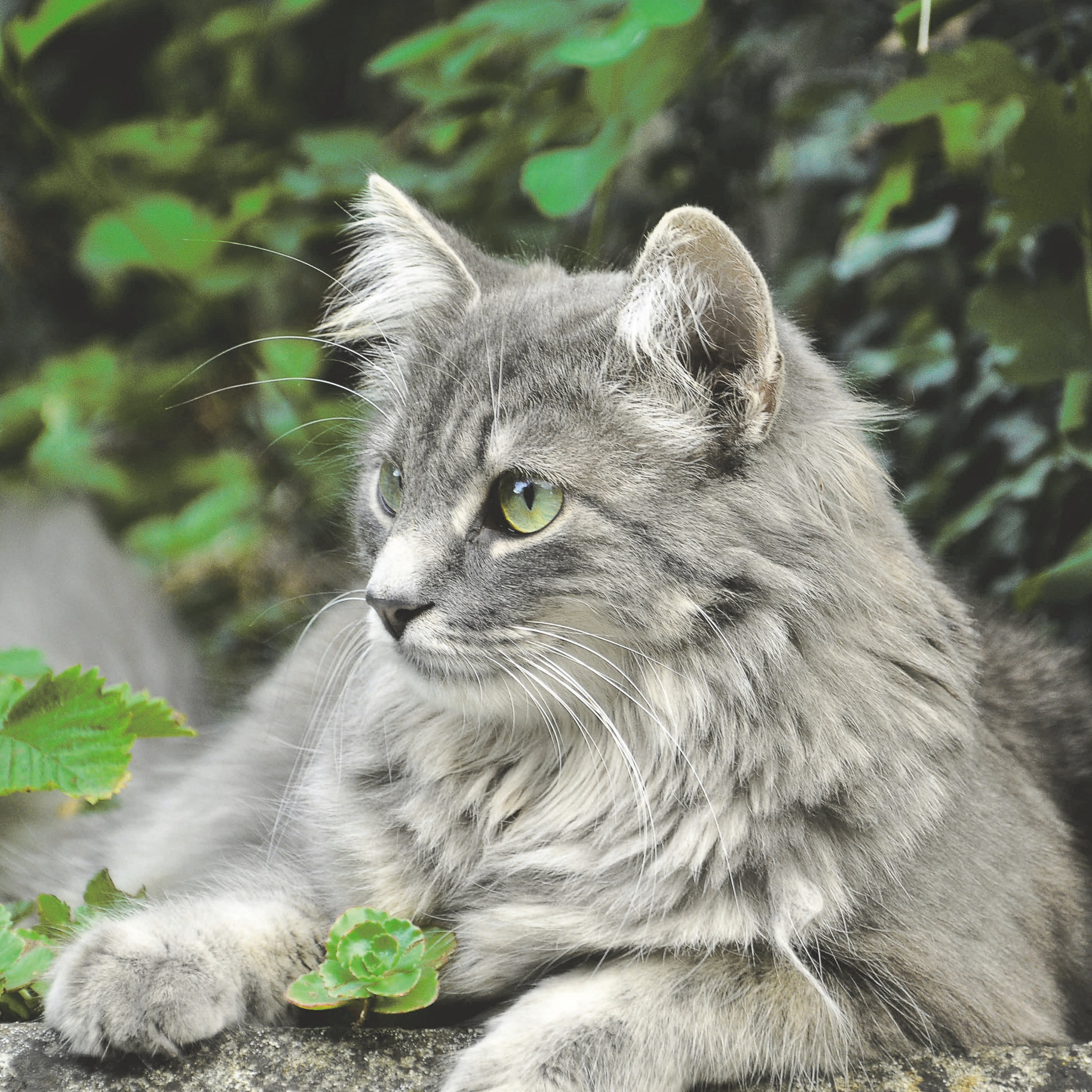}
			}; \\
		}; 
	\end{tikzpicture}
};

\node[imgbox2, anchor=north west] (step1block) at ([xshift=2.2cm]img.north east){
	\begin{tikzpicture}
	\def\textx{0} 
	\def\imgx{0.6}     
	
	\node[anchor=west,font=\Large] (theta) at (\textx,0) {\;$\theta_t$};
	\node[anchor=west] (noise_theta) at (\imgx,0) {\includegraphics[width=1cm]{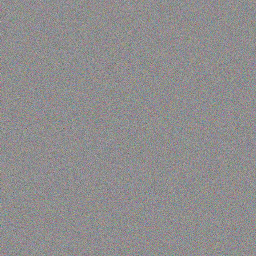}};
	
	\node[anchor=west,font=\Large] (mt) at (\textx,-1.1) {\;$m_t$};
	\node[anchor=west] (noise_m) at (\imgx,-1.1) {\includegraphics[width=1cm]{figures/noise.pdf}};

	\path (noise_theta.north east) -- (noise_m.south east) coordinate[pos=0.38] (midpoint);
		 \node[draw,densely dashed,very thick,sharp corners,inner sep=2pt] (thetahat) at ([xshift=2pt]midpoint)
		{$\tilde{\theta}_t = (1 - \alpha_t)\theta_t + \alpha_t m_t$};
		\draw[-stealth,very thick] ([yshift=6pt]noise_theta.east) to[out=0, in=110] ([xshift=30pt] thetahat.north west);
		\draw[-stealth,very thick] ([yshift=-6pt]noise_m.east)    to[out=0, in=250] ([xshift=30pt] thetahat.south west);
	
	\node[
	rotate=-90,
	transform shape,
	anchor=center,
	text=red,
	font=\bfseries
	]
	at ([xshift=10pt]thetahat.east) {Step 1};

	\end{tikzpicture}
};

\draw[very thick] 
([xshift=-0.6cm]step1block.east |- step1block.north) --
([xshift=-0.6cm]step1block.east |- step1block.south);

\begin{pgfonlayer}{background}
	\coordinate (bb-sw) at ([xshift=-0.6cm]step1block.south east);
	\coordinate (bb-ne) at (step1block.north east);
	\begin{scope}
		\path[clip, rounded corners=4pt]
		(step1block.south west) rectangle (step1block.north east);
		
		\path[clip] (bb-sw) rectangle (bb-ne);
		
		\fill[LightGrey!20] (bb-sw) rectangle (bb-ne);

	\end{scope}
\end{pgfonlayer}

\coordinate (img_topright) at ([xshift=0pt,yshift=-10pt] img.north east);
\coordinate (mt_pos)    at ([xshift=7pt, yshift=-10pt] step1block.west);
\coordinate (branchpoint) at (img_topright);
\coordinate (theta_pos) at (mt_pos |- branchpoint);

\draw[-stealth, very thick] (img_topright) --  node[pos=0.5, above,align=center,inner sep=1pt] {Initialization\\[-0.6ex]$\theta_0$, $m_0$} (theta_pos.west);
\coordinate (branchpoint) at ([xshift=2.0cm] img_topright);
\draw[-stealth,very thick] (branchpoint) |- (mt_pos.west);

\node[imgbox_horizontal_narrow, anchor=north west] at ([yshift=-1.6cm]img.south west) (step2block) {
	\begin{tikzpicture}[baseline=(current bounding box.center)]
		
	 \matrix[
	ampersand replacement=\&, column sep=2pt, row sep=0pt,
	nodes={anchor=center}
	] (m) {
		\node[
		text=red,
		font=\bfseries,
		inner sep=0pt,
		outer sep=0pt,      
		yshift=-0.8cm,
		] (step2label) {
			Step 2 \\[0.5em]
			\normalfont\textcolor{black}{Gradient}\\[0.5em]
			\normalfont\textcolor{black}{Estimation}
		}; \&
		 \node[anchor=center, align=center,inner sep=0pt,font=\small] (g1node){
		 	\begingroup
		 	\setlength{\jot}{1pt}
		 	$
		 	\begin{aligned}
		 		 g_1(\tilde{\theta}_t) &= \frac{1}{\epsilon} \left(f(\tilde{\theta}_t \!+\! \epsilon \bv_t) - f(\tilde{\theta}_t)\right) \bv_t \\
		 		&+ \textstyle\sum_{i=1}^s \frac{1}{\epsilon} \left(f(\tilde{\theta}_t \!+\! \epsilon \bp_{t,i}) - f(\tilde{\theta}_t)\right) \bp_{t,i}
		 	\end{aligned}$
		 	\endgroup
		 };
		 
		 \node[anchor=center, below=0.1cm of g1node,align=center, inner sep=0pt,font=\small] (g2node) {
		 	\begingroup
		 	\setlength{\jot}{1pt}
		 	$
		 	\begin{aligned}
		 		 g_2(\tilde{\theta}_t) &= \frac{(d\!-\!s)\pi}{2\epsilon(q\!-\!s\!-\!1) + \pi\epsilon}  \left(f(\tilde{\theta}_t \!+\! \epsilon \bv_t) - f(\tilde{\theta}_t)\right) \bv_t \\[0.1pt]
		 		&+ \textstyle\sum_{i=1}^s \frac{1}{\epsilon} \left(f(\tilde{\theta}_t \!+\! \epsilon \bp_{t,i}) - f(\tilde{\theta}_t)\right) \bp_{t,i}
		 	\end{aligned}$
		 	\endgroup
		 };

		 \draw[densely dashed, thick] (g2node.north west) -- (g2node.north east);
		 \path let \p1 = (g2node.east) in
		 coordinate (rightbarx) at (\x1,0);
		 \draw[densely dashed, thick]
		 (rightbarx |- g1node.north) -- (rightbarx |- g2node.south);
		 
		 \path (g1node.east) -- (g2node.east) coordinate[pos=0.5] (midpoint);

		\node[anchor=west,sharp corners,inner sep=0pt, outer sep=0pt] (illustration_a) 
		at ([xshift=10pt]midpoint)  {
		 	\begin{tikzpicture}[scale=0.7]

		 		\def\r{1}
		 		\def\delta{5}   
		 		\coordinate (start_curve) at (2.3,4.6);
		 		\coordinate (lowest_point) at (3.8,3.4);
		 		\coordinate (end_curve) at (4.5,4.6);

		 		\shade[
		 		shading = radial,
		 		inner color = white,      
		 		outer color = NavyBlue,    
		 		shading angle=0,
		 		opacity = 0.9             
		 		]
		 		(start_curve) parabola bend (lowest_point) (end_curve);

		 		\draw[very thick] (start_curve) parabola bend (lowest_point) (end_curve);
		 		\path[name path=curve](start_curve) parabola bend (lowest_point) (end_curve);

		 		\pgfresetboundingbox \path (start_curve); \path (end_curve); 

		 		\coordinate (O) at  (3.2,2.1);

		 		\path[name path=mycircle]
		 		(O) ++(180+\delta:\r)    
		 		arc[
		 		start angle=180+\delta,
		 		end angle=-10,  
		 		radius=\r
		 		];
		 		
		 		\coordinate (theta_t) at (2.7,4);
		 		\coordinate (tilde_theta_t) at (3,4);
		 		\coordinate (theta_t_plus_1) at (3.5,4);
		 		\coordinate (m_t) at (4.7,4);
		 		\coordinate (m_t_plus_1) at (5.1,4);

		 		\path[name path=ray_theta_t] (O) -- ($(O)!1.1!(theta_t)$);
		 		\path[name intersections={of=ray_theta_t and curve, by=I_theta_t}];
		 		\path[name intersections={of=mycircle and ray_theta_t, by=IC_theta_t}];

		 		\path[name path=ray_tilde_theta_t] (O) -- (tilde_theta_t);
		 		\path[name intersections={of=ray_tilde_theta_t and curve, by=I_tilde_theta_t}];
		 		
		 		\coordinate (g1) at ($ (I_tilde_theta_t) !{\r cm}! ($(I_tilde_theta_t)+(-0.4,-0.2)$) $);
		 		\draw[-stealth] (I_tilde_theta_t) -- (g1) node[anchor=south,yshift=4pt,font=\small]{$g_1(\tilde{\theta}_t)$};

		 		\coordinate (g2) at ($ (I_tilde_theta_t) !{1.22\r cm}! ($(I_tilde_theta_t) + (-0.7, -0.6)$) $);
		 		\draw[-stealth] (I_tilde_theta_t) -- (g2) node[anchor=north,font=\small]{$g_2(\tilde{\theta}_t)$};

		 		\path[name intersections={of=mycircle and ray_tilde_theta_t, by=IC_tilde_theta_t}];

		 		\draw[-stealth] (O) -- (IC_tilde_theta_t) node[anchor=north west,xshift=2pt,yshift=-2pt,font=\small]{$\tilde{\theta}_t$};		 		
		 		\draw[densely dashed] (IC_tilde_theta_t) -- (I_tilde_theta_t) node[anchor=south west, font=\small,color=black,xshift=2pt]{$f(\tilde{\theta}_t)$};
		 		\fill[orange] (I_tilde_theta_t) circle (2pt);

		 		\coordinate (bb-sw) at (current bounding box.south west);
		 		\coordinate (bb-ne) at (current bounding box.north east);
		 		 \begin{scope}[on background layer]
		 			\fill[RegionYellow] (bb-sw) rectangle (bb-ne);
		 		\end{scope}
		 		\draw[very thick] (bb-sw) rectangle (bb-ne);
		 		
		 		\fill[fill=black] (O) circle[radius=2pt];
		 		\begin{scope}
		 			\clip (bb-sw) rectangle (bb-ne);
		 			\draw[densely dashed] (O) circle (\r);
		 		\end{scope}

		 	\end{tikzpicture}
		 	
		 };
		 \node[draw=black, thick, densely dashed, rounded corners=4pt, inner sep=1pt, fit=(g1node)(g2node)(illustration_a)] (fullbox) {};
		 
		 \node[anchor=north west] (step2_rightblock)
		 at (fullbox.north east |- fullbox.north) [xshift=15pt] {
		 	\begin{tikzpicture}[baseline=(current bounding box.north)]
		 		\node[align=center, font=\scriptsize] {
		 			$\displaystyle\bv_t = \frac{1}{\sqrt{q\!-\!s}} \sum_{i=1}^{q-s} \underbrace{\operatorname{sign}\left(f(\tilde{\theta}_t \!+\! \epsilon \bu_i)\!-\! f(\tilde{\theta}_t)\right)}_{\text{\scriptsize only a single query (Eq. \eqref{eq:sign_opt_single_query})}}\bu_i$
		 		};
		 	\end{tikzpicture}
		 };
		 \node[anchor=north,sharp corners] (illustration_b)  at (step2_rightblock.south) {
		 	\begin{tikzpicture}[scale=0.7]
		 		\def\r{1}
		 		\def\delta{5}
		 		\coordinate (start_curve) at (2.3,4.6);
		 		\coordinate (lowest_point) at (3.8,3.4);
		 		\coordinate (end_curve) at (5,4.6);
		 		
		 		\shade[shading=radial, inner color=white, outer color=NavyBlue, opacity=0.9]
		 		(start_curve) parabola bend (lowest_point) (end_curve);
		 		
		 		\draw[very thick] (start_curve) parabola bend (lowest_point) (end_curve);
		 		\path[name path=curve](start_curve) parabola bend (lowest_point) (end_curve);
		 		
		 		\pgfresetboundingbox \path (start_curve); \path (end_curve); 
		 		
		 		\coordinate (O) at  (3.2,2.65);
		 		
		 		\path[name path=mycircle]
		 		(O) ++(180+\delta:\r) arc[start angle=180+\delta, end angle=-10, radius=\r];
		 		
		 		\coordinate (tilde_theta_t) at (3,4);
		 		\path[name path=ray_tilde_theta_t] (O) -- (tilde_theta_t);
		 		\path[name intersections={of=ray_tilde_theta_t and curve, by=I_tilde_theta_t}];
		 		\path[name intersections={of=mycircle and ray_tilde_theta_t, by=IC_tilde_theta_t}];
		 		
		 		\draw[-stealth] (O) -- ($(O)!0.94!(I_tilde_theta_t)$) node[anchor=north west,xshift=2pt,yshift=-4pt,font=\small]{$\tilde{\theta}_t$};

			 	\def\uradius{0.6} 
			 	\foreach \i/\angle/\xshift/\yshift in {
			 		1/10/6pt/2pt,
			 		2/50/4pt/2pt,
			 		3/100/0pt/2pt,
			 		4/150/-6pt/3pt,
			 		5/200/-6pt/1pt
			 	} {
			 		\path (I_tilde_theta_t) ++(\angle:\uradius) coordinate (P\i);
			 		\draw[-stealth] (I_tilde_theta_t) -- (P\i)
			 		node[pos=1, font=\small,  xshift=\xshift, yshift=\yshift] {$\epsilon\mathbf{u}_{\i}$};
			 		
			 	}
	
		 		\fill[orange] (I_tilde_theta_t) circle (2pt);
		 		
		 		\coordinate (bb-sw) at (current bounding box.south west);
		 		\coordinate (bb-ne) at (current bounding box.north east);
		 		\begin{scope}[on background layer]
		 			\fill[RegionYellow] (bb-sw) rectangle (bb-ne);
		 		\end{scope}
		 		\draw[very thick] (bb-sw) rectangle (bb-ne);
		 		
		 		\fill[black] (O) circle[radius=2pt];
	
		 	\end{tikzpicture}
		 };
		 \coordinate (fullbox-top) at ([xshift=1cm]fullbox.north east);
		 \coordinate (fullbox-bottom) at ([xshift=1cm]fullbox.south east);
		 \node[draw=black, thick, densely dashed, rounded corners=4pt, inner sep=0pt,
		 fit=(step2_rightblock)(illustration_b)(fullbox-top)(fullbox-bottom)] 
		 (fullbox2) {};
		  \draw[-stealth, very thick] 
		  (fullbox2.west) -- node[midway, above,font=\Large] {$\bv_t$} 
		  (fullbox.east);\\
		  	};

	\end{tikzpicture}
};

\begin{pgfonlayer}{background}
	\coordinate (bb-sw) at (step2block.south west);
	\coordinate (bb-ne) at (step2block.north east);
	\path let \p1 = (bb-sw), \p2 = (bb-ne) in coordinate (split-x) at ({\x1 + 1.65cm}, 0);
	
	\begin{scope}
		\path[clip, rounded corners=4pt] (bb-sw) rectangle (bb-ne);
		\fill[LightGrey!20] (bb-sw) rectangle (split-x |- bb-ne);
		\draw[very thick] 
		(split-x |- bb-ne) -- (split-x |- bb-sw);
	\end{scope}
\end{pgfonlayer}

\path let
\p1 = (img.north east),         
\p2 = (current bounding box.east)  
in coordinate (target_point) at (\x2, \y1); 
\node[imgbox, anchor=north east] 
at (target_point) (random_directions_block) {
	\begin{tikzpicture}
	 \node (graphic) at (2,0) {
	\begin{tikzpicture}[baseline=(current bounding box.center)]
		\coordinate (O) at (1, 1); 
		\foreach \i/\x/\y/\xshift/\yshift in {
		  1/0.59750776/1.80598447/0/0,
		2/1.63639610/1.63639610/6/6,
		3/1.9/1.0/11/2,
		q-s/1.40249224/0.19401553/17/0
		}{
			\coordinate (P\i) at (\x,\y);
			\draw[-stealth, very thick] (O) -- (P\i) node[pos=1.05, xshift=\xshift,yshift=\yshift] {$\mathbf{u}_{\i}$};
		}
		
		\coordinate (orig_img) at (0,0.6);
		\filldraw[black] (orig_img) circle (2pt);
		
		\node at (2,0.6) {\dots};
		\draw[-stealth,very thick,color=NavyBlue] (O) -- (0.38994905, 0.33830684) node[anchor=north] {$\mathbf{p}_{t,1}$};
		\node[color=NavyBlue] at (1,0.5) {\rotatebox{-30}{\dots}};
		\draw[-stealth,very thick,color=NavyBlue] (O) -- (1, 0.1) node[anchor=north,yshift=-0.6pt] {$\mathbf{p}_{t,s}$};
		\draw[-stealth,very thick] (orig_img) -- ($(orig_img)!0.94!(O)$) node[pos=0, anchor=north east] {$\mathbf{x}$} node[pos=0.4, anchor=south] {$\tilde{\theta}_t$};;
	
		\fill (O) circle[radius=2pt];

	\end{tikzpicture}
	};
	\node[anchor=west, font=\small, align=left]
	at ([xshift=-4cm]graphic.center)  
	{
		Sample $q-s$ random\\vectors, take $s$ priors,\\%
		then orthogonalize\\via Gram–Schmidt.
	};
	
\end{tikzpicture}
};

\node[imgbox_all_narrow,anchor=north west,color=StepBoxPink] at ([yshift=-0.3cm]img.south west) (step3block) {

	\begin{tikzpicture}
		\matrix[
		ampersand replacement=\&,
		column sep=2pt,
		row sep=0pt, nodes={anchor=center}  
		](m) {
		\node[
		align=center,
		text=red,
		font=\bfseries,
		inner sep=0pt,
		minimum width=1.65cm
		] (step3label) {Step 3\\\normalfont\textcolor{black}{Update}}; 
		\&
			\node[
			align=center,
			font=\small,
			inner sep=1pt
			] (step3formula) {
	
				$
				\begin{aligned}
					\theta_{t+1} &= \tilde{\theta}_t - 1/\hat{L} \cdot g_1(\tilde{\theta}_t) \\[-4pt]
					m_{t+1} &= m_t - \zeta_t/\alpha_t \cdot g_2(\tilde{\theta}_t)
				\end{aligned}
				$

			}; \\
		};

	\end{tikzpicture}
};

\begin{pgfonlayer}{background}
	\coordinate (bb-sw) at (step3block.south west);  
	\coordinate (bb-ne) at (step3block.north east);  
	\path let \p1 = (bb-sw), \p2 = (bb-ne) in coordinate (split-x) at ({\x1 + 1.65cm}, 0);
	\begin{scope}
		\path[clip, rounded corners=4pt] (bb-sw) rectangle (bb-ne);
		\fill[StepBoxPinkLight, opacity=0.2] (bb-sw) rectangle (split-x |- bb-ne);
		\draw[StepBoxPink, very thick] 
		(split-x |- bb-ne) -- (split-x |- bb-sw);
	\end{scope}
\end{pgfonlayer}

	\coordinate (img_topleft) at ([xshift=4cm,yshift=-4pt] step2block.north west);
	\coordinate (img_topleft_bottom) at ([xshift=4cm] step3block.south west);
	\coordinate (img_topleft_top) at ([xshift=4cm] step3block.north west);
	\coordinate (branchpoint) at (img_topright);
	\coordinate (mt_pos)    at ([xshift=7pt, yshift=-50pt] step1block.north west);
	\coordinate (theta_pos) at ([xshift=7pt, yshift=-20pt] step1block.north west);

	\draw[very thick,color=StepBoxPink] (img_topleft) -- (img_topleft_bottom);
	\draw[-stealth,very thick,color=StepBoxPink] (img_topleft_top) |- node[pos=0.5, below right,font=\normalfont,yshift=2pt] {$g_1(\tilde{\theta}_t)$} (theta_pos.west);
	\draw[-stealth, very thick, color=StepBoxPink] 
	(img_topleft_top |- mt_pos) -- node[pos=0, above right,font=\normalfont,yshift=-3pt] {$g_2(\tilde{\theta}_t)$} (mt_pos.west);

	\coordinate (step1_bottom_right) at ([xshift=-2cm] step1block.south east);
	\coordinate (step2_top_right) at ([xshift=-2cm,yshift=0.4cm] step2block.north east);
	\coordinate (step2_top_right_) at ([xshift=-2cm] step2block.north east);
	\draw[very thick] (step1_bottom_right) |- node[pos=0.25, anchor=west,font=\Large] {$\tilde{\theta}_t$} (step2_top_right);
	\draw[-stealth, very thick]  (step2_top_right |- random_directions_block.south) -- node[pos=0.5,anchor=west,font=\Large,align=center]{$\{\bu_i\}_{i=1}^{q-s}$\\$\{\bp_{t,i}\}_{i=1}^{s}$} (step2_top_right_);

	\end{tikzpicture}
	\caption{Illustration of one iteration in PARS-OPT. We first form a lookahead point $\tilde{\theta}_t$ by linearly interpolating between the current direction $\theta_t$ and the momentum term $m_t$ (with $m_0 = \theta_0$). Next, we estimate $\bv_t$ via a sign-based procedure over a set of randomly sampled orthonormal basis vectors. Finally, we use $\bv_t$ to compute the biased gradient estimate $g_1(\tilde{\theta}_t)$ and the unbiased estimate $g_2(\tilde{\theta}_t)$, which are then used to update $\theta_t$ and $m_t$, yielding $\theta_{t+1}$ and $m_{t+1}$ for the next iteration.}
		\label{fig:algorithm}
\end{figure*}

\subsection{ARS-OPT}
Our framework, spanning from Step \ref{step:perturb} to Step \ref{step:update}, is compatible with various gradient estimation techniques, enabling flexible algorithmic implementations. In this section, we provide a detailed introduction to the fundamental algorithm, ARS-OPT.
In Step \ref{step:perturb}, unlike standard gradient descent, the gradient is not computed at the current direction $\theta_t$. Instead, the algorithm predicts a candidate ray direction $\tilde{\theta}_t$ by interpolating between the momentum vector $m_t$ and the current direction $\theta_t$. 
The sequences of $\theta_t$ and $m_t$ are referred to as \emph{the main sequence} and \emph{the auxiliary sequence}, respectively.
$\tilde{\theta}_t$ is referred to as the \emph{lookahead position} of $\theta_t$, and is computed via interpolation: $\tilde{\theta}_t \gets (1 - \alpha_t)\theta_t + \alpha_t m_t$, where $\alpha_t \in [0,1]$ is the interpolation coefficient. The value of $\alpha_t$ is defined as the positive root of the equation $\alpha_t^2 = \zeta_t \gamma_t (1-\alpha_t)$, where $\gamma_t$ is a scalar determined in Algorithm \ref{alg:pars-opt}, and $\zeta_t = \left(\frac{2(q-1)+\pi}{d\pi}\right)/ \left(\hat{L} \left(\frac{d\pi}{2(q-1)+\pi}\right)\right)$. This expression is derived from the convergence analysis of ARS-OPT. This choice of $\alpha_t$ is critical to establishing the algorithm's theoretical convergence guarantees. For detailed derivations, we refer readers to Appendix A. To maintain two sequences---the optimization variable $\theta_t$ and the auxiliary variable $m_t$ (which accumulates historical momentum to capture global optimization trends)---we employ two gradient estimates, $g_1(\tilde{\theta}_t)$ and $g_2(\tilde{\theta}_t)$, to update $\theta_t$ and $m_t$, respectively:
\begin{footnotesize}
    \begin{align}
        g_1(\tilde{\theta}_t) &\coloneqq \nabla f(\tilde{\theta}_t)^\top \bv_t \cdot \bv_t \approx \frac{f(\tilde{\theta}_t + \epsilon \bv_t) - f(\tilde{\theta}_t)}{\epsilon}\cdot\bv_t, \label{eq:ARS_g1} \\ 
        g_2(\tilde{\theta}_t) &\coloneqq \frac{d}{\frac{2}{\pi}(q-1)+1} \nabla f(\tilde{\theta}_t)^\top \bv_t \cdot \bv_t \label{eq:ARS_g2} \\
        &\approx \frac{d\left(f(\tilde{\theta}_t + \epsilon \bv_t) - f(\tilde{\theta}_t)\right)}{\frac{2\epsilon}{\pi}(q-1)+\epsilon}\cdot \bv_t, \label{eq:ARS_g2_fd}
    \end{align}
\end{footnotesize}
where $d$ is the dimension of the input image, $q$ is the number of vectors in gradient estimation, and $\bv_t$ is the sign-based gradient estimate \cite{cheng2020sign} as $\bv_t \coloneqq \frac{1}{\sqrt{q}}\sum_{i=1}^{q}\text{sign}(f(\tilde{\theta}_t+\epsilon \bu_i)-f(\tilde{\theta}_t))\bu_i$, which calculates the sign of the directional derivative with a single query: 
\begin{footnotesize}
    \begin{equation}
    \label{eq:sign_opt_single_query}
    \text{sign}(f(\theta + \epsilon\bu)- f(\theta)) = \begin{cases} +1, & \Phi\left(\bx + f(\theta) \frac{\theta + \epsilon \bu}{\|\theta + \epsilon \bu\|}\right) \ne 1 \\ -1, & \text{otherwise}\end{cases}.
    \end{equation}
\end{footnotesize}
Eq. \eqref{eq:ARS_g1} can be regarded as the projection of the true gradient onto $\bv_t$. Eq. \eqref{eq:ARS_g2} is an unbiased estimator of $\nabla f(\tilde{\theta}_t)$, derived from Theorem \ref{theorem:proportional_estimator}\footnote{Throughout this paper, for any vector $\bv$, we denote $\overline{\bv}$ as its $\ell_2$-normalized vector, where $\overline{\bv}\coloneqq \frac{\bv}{\|\bv\|}$.}.
\begin{theorem}
\label{theorem:proportional_estimator}
Let $\{\bu_1, \bu_2, \dots,\bu_q\}$ be an orthonormal set obtained by orthogonalizing $q$ vectors independently and uniformly sampled from the unit
sphere in $\mathbb{R}^d$.  Suppose $\bg$ is a fixed vector in $\mathbb{R}^d$ (for example, it is the true gradient to be estimated). Let $\bv\coloneqq\sum_{i=1}^q \operatorname{sign} (\bg^\top \bu_i) \bu_i$, and $\hat{\bg}\coloneqq \bg^\top \overline{\bv}\cdot \overline{\bv}$. Then we have 
\begin{footnotesize}
    \begin{equation}
    \E[\hat{\bg}] = \E[(\overline{\bg}^\top \overline{\bv})^2] \cdot \bg. \label{eq:unbiased_estimator_g2}
    \end{equation}
\end{footnotesize}
\end{theorem}
The proof of Theorem \ref{theorem:proportional_estimator} is shown in Appendix A. In Eq.~\eqref{eq:unbiased_estimator_g2}, $\hat{\bg}$ is equal to $g_1(\tilde{\theta}_t)$, and $\E[(\overline{\bg}^\top \overline{\bv})^2]=\frac{1}{d}\left( \frac{2}{\pi}(q-1)+1 \right)$
 based on Lemma A.5 (see Appendix A). Thus we have $\E[g_1(\tilde{\theta}_t)] = \frac{1}{d}\left( \frac{2}{\pi}(q-1)+1 \right) \cdot \bg$. Consequently, the true gradient can be recovered as $\bg =  \frac{d}{\frac{2}{\pi}(q-1)+1} \E[g_1(\tilde{\theta}_t)] = \frac{d}{\frac{2}{\pi}(q-1)+1} \E[\nabla f(\tilde{\theta}_t)^\top \bv_t \cdot \bv_t]$, which shows that $g_2(\tilde{\theta}_t)$ is an unbiased estimator of $\nabla f(\tilde{\theta}_t)$.

\subsection{PARS-OPT}
\label{sec:PARS-OPT}

ARS-OPT relies exclusively on random orthonormal vectors to estimate the gradient, which leads to inaccurate gradient approximation and poor query efficiency. To further enhance the efficiency of the algorithm, we propose a variant algorithm named Prior-guided ARS-OPT (PARS-OPT) within our framework. An ideal prior would be the gradient of $\hat{f}(\theta)$ derived from a surrogate model. However, since $\hat{f}(\theta)$ is non-differentiable due to the binary search process, this gradient cannot be directly computed. To overcome this challenge, we employ a differentiable surrogate function $h(\theta, \lambda)$ in Eq.~\eqref{eq:surrogate_loss}, following Ma et al. \citeyearpar{ma2025boosting}, which ensures the gradient relationship: $\nabla \hat{f}(\theta_0) = c \cdot \nabla_\theta h(\theta_0, \lambda_0)$ for any non-zero vector $\theta_0 \in \mathbb{R}^d$ with $\hat{f}(\theta_0) < +\infty$. Here, $\hat{f}(\cdot)$ is defined on the surrogate model $\hat{\psi}$, $\lambda_0 = \hat{f}(\theta_0)$ is treated as a constant scalar during differentiation, and $c$ is a non-zero constant.
\begin{small}
\begin{align}
    \label{eq:surrogate_loss}
    h(\theta, \lambda) \coloneqq \begin{cases}
        \hat{\psi}_y - \max_{j\neq y} \hat{\psi}_j, & \text{\small{if untargeted attack,}} \\
        \max_{j\neq y_\text{adv}} \hat{\psi}_j - \hat{\psi}_{y_\text{adv}}, & \text{\small{if targeted attack,}} \\
    \end{cases}
\end{align}
\end{small}
where $\hat{\psi}_i \coloneqq \hat{\psi}\big(\mathbf{x} + \lambda \cdot \frac{\theta}{\|\theta\|}\big)_i$ is an abbreviation for the $i$-th element of the output of the surrogate model $\hat{\psi}$, and $\mathbf{x}$ is the original image. Given $s$ non-zero vectors $\mathbf{k}_{t,1}, \dots, \mathbf{k}_{t,s}$ computed as $\nabla_\theta h(\theta_0, \lambda_0)$ from $s$ surrogate models and $q - s$ randomly sampled vectors $\mathbf{r}_1,\dots,\mathbf{r}_{q-s} \sim \mathcal{N}(\mathbf{0}, \mathbf{I})$, we apply Gram-Schmidt orthogonalization to these $q$ vectors to obtain an orthonormal set $\mathbf{p}_{t,1}, \dots, \mathbf{p}_{t,s}, \mathbf{u}_1, \dots, \mathbf{u}_{q-s}$, which are used by the gradient estimation formulas:
{\fontsize{7pt}{9pt}\selectfont
\begin{align}
    &g_1(\tilde{\theta}_t) = \nabla f(\tilde{\theta}_t)^\top \bv_t \cdot \bv_t + \sum\limits_{i=1}^s \nabla f(\tilde{\theta}_t)^\top \bp_{t,i} \cdot \bp_{t,i} \label{eq:pars_g1}\\
	&\approx \frac{f(\tilde{\theta}_t + \epsilon \bv_t) - f(\tilde{\theta}_t)}{\epsilon}\bv_t + \sum_{i=1}^s \frac{f(\tilde{\theta}_t + \epsilon \bp_{t,i}) - f(\tilde{\theta}_t)}{\epsilon} \bp_{t,i}. \label{eq:pars_g1_fd} \\
    &g_2(\tilde{\theta}_t) = \frac{d-s}{\frac{2}{\pi}(q-s-1)+1} \nabla f(\tilde{\theta}_t)^\top \bv_t \cdot \bv_t + \sum\limits_{i=1}^s \nabla f(\tilde{\theta}_t)^\top \bp_{t,i} \cdot \bp_{t,i} \label{eq:pars_g2}\\
    &\approx \frac{(d-s)\left(f(\tilde{\theta}_t + \epsilon \bv_t) - f(\tilde{\theta}_t)\right)}{\frac{2\epsilon}{\pi}(q-s-1)+\epsilon}\bv_t + \sum_{i=1}^s \frac{f(\tilde{\theta}_t + \epsilon \bp_{t,i}) - f(\tilde{\theta}_t)}{\epsilon}\bp_{t,i}, \label{eq:pars_g2_fd}
\end{align}}
where $\bv_t \coloneqq \frac{1}{\sqrt{q-s}}\sum_{i=1}^{q-s}\text{sign}(f(\tilde{\theta}_t+\epsilon \bu_i)-f(\tilde{\theta}_t))\bu_i$. To ensure the convergence of PARS-OPT, we still require $g_2(\tilde{\theta}_t)$ to be an unbiased estimator of $\nabla f(\tilde{\theta}_t)$, whose proof is more involved than in ARS-OPT; see the Appendix for details.


\begin{algorithm}[!t]
\caption{(P)ARS-OPT Attack}
\label{alg:pars-opt}
\begin{algorithmic}[1]
     \STATE \textbf{Input:} $L$-smooth function $f$, $\hat{L}\geq L$, the original image $\bx$, the success indicator function $\Phi(\cdot)$, initial ray direction $\theta_0$, number of estimation vectors $q$, finite-difference step size $\epsilon$, input dimension $d$, number of iterations $T$, maximum gradient norm $g_{\max}$, $\gamma_0>0$, surrogate model set $\mathbb{S} = \{\hat{\psi}^{(i)}, \dots, \hat{\psi}^{(s)}\}$ with $s > 0$ for PARS-OPT, and $\mathbb{S} = \emptyset$ for ARS-OPT.
    \STATE \textbf{Output:} Adversarial example $\bx_\text{adv}$.
    \STATE $m_0 \gets \theta_0$,\quad $\|\hat{\nabla} f_{-1} \|^2 \gets +\infty$;
    \FOR{$t = 0$ to $T-1$}
    \FOR{$\hat{\psi}^{(i)}$ {\bfseries{in}} $\mathbb{S}$}
    \STATE $\lambda_{t} \gets \text{BinarySearch}(\mathbf{x}, \theta_t, \hat{\psi}^{(i)}, \Phi)$;
    \STATE $\mathbf{k}_{t,i}\gets$ $\nabla_\theta h(\theta_t, \lambda_t)$ on $\hat{\psi}^{(i)}$ with $\lambda_t$ treated as a constant in differentiation; \COMMENT{obtain $s$ priors.}
    \ENDFOR
    \STATE $\mathbf{r}_{i} \sim \mathcal{N}(\mathbf{0}, \mathbf{I})$ for $i=1,\dots,q-s$;
    \STATE $\mathbf{p}_{t,1},\dots,\mathbf{p}_{t,s}, \mathbf{u}_1,\dots,\mathbf{u}_{q-s} \gets$\\ Orthogonalize($\{\mathbf{k}_{t,1},\dots,\mathbf{k}_{t,s}, \mathbf{r}_{1},\dots,\mathbf{r}_{q-s}\}$);
    \STATE $\hat{D}_t \gets \frac{\sum_{i=1}^s(\nabla f(\theta_{t})^\top \bp_{t,i})^2}{\| \hat{\nabla} f_{t-1} \|^2}$; \COMMENT{It requires extra queries.}
    \STATE $\zeta_t \gets \frac{\hat{D}_t + \frac{(2(q-s-1)+\pi)}{(d-s)\pi}(1-\hat{D}_t)}{\hat{L}\left(\hat{D}_t + \frac{(d-s)\pi}{(2(q-s-1)+\pi)}(1-\hat{D}_t)\right)}$;
    \STATE $\tilde{\theta}_t \gets (1 - \alpha_t)\theta_t + \alpha_t m_t$, where $\alpha_t \geq 0$ is a positive root of the equation $\alpha_t^2 = \zeta_t\gamma_t(1-\alpha_t)$; 
    \STATE $\gamma_{t+1} \gets (1-\alpha_t)\gamma_t$;
    \STATE $\bv_t \gets \frac{1}{\sqrt{q-s}}\sum_{i=1}^{q-s} \text{sign}(f(\tilde{\theta}_t + \epsilon \bu_i) - f(\tilde{\theta}_t)) \bu_i$;
    \STATE $\nabla f(\tilde{\theta}_t)^\top \bv_t \gets \frac{f(\tilde{\theta}_t + \epsilon \bv_t) - f(\tilde{\theta}_t)}{\epsilon}$; \COMMENT{Directional derivative approximation by finite differences.}
    \STATE $\{\nabla f(\tilde{\theta}_t)^\top \bp_{t,i}\}_{i=1}^s
    \gets \left\{\frac{f(\tilde{\theta}_t + \epsilon \bp_{t,i}) - f(\tilde{\theta}_t)}{\epsilon}\right\}_{i=1}^s$;
    \STATE Estimate $g_1(\tilde{\theta}_t)$, $g_2(\tilde{\theta}_t)$ by using Eq. \eqref{eq:pars_g1_fd} and Eq. \eqref{eq:pars_g2_fd};
    \STATE $g_1(\tilde{\theta}_t) \gets \text{ClipGradNorm} \big(g_1(\tilde{\theta}_t), g_{\max}\big)$;
    \STATE 
    \begin{scriptsize}
        $\| \hat{\nabla} f_t \|^2 \gets \sum\limits_{i=1}^s \left(\nabla f(\tilde{\theta}_t)^\top \bp_{t,i}\right)^2 + \frac{(d-s)\pi}{2(q-s-1)+\pi} \left(\nabla f(\tilde{\theta}_t)^\top \bv_t\right)^2$; \\
    \end{scriptsize}
     \COMMENT{This line is used only in PARS-OPT.}
     \STATE $\theta_{t+1} \gets \tilde{\theta}_t - \frac{1}{\hat{L}} g_1(\tilde{\theta}_t)$, \quad $m_{t+1} \gets m_t - \frac{\zeta_t}{\alpha_t}g_2(\tilde{\theta}_t)$;
     
    \ENDFOR
    \STATE\textbf{return} $\bx_\text{adv} \gets \bx + f(\theta_T)\frac{\theta_T}{\| \theta_T \|}$.
    \end{algorithmic}
\end{algorithm}

Algorithm \ref{alg:pars-opt} presents a unified framework covering both ARS-OPT and PARS-OPT, and Fig. \ref{fig:algorithm} offers an overview of the PARS-OPT procedure. In targeted attacks, we initialize $\theta_0$ with the direction to an image $\tilde{\mathbf{x}}_0$ from the target class in the training set. The momentum term $m_0$ is initialized as $\theta_0$ in the first iteration. Specifically, setting $s = 0$ reduces Eq.~\eqref{eq:pars_g1_fd} and Eq. \eqref{eq:pars_g2_fd} to their counterparts in ARS-OPT, namely Eq.~\eqref{eq:ARS_g1} and Eq. \eqref{eq:ARS_g2_fd}. 
Note that $\hat{D}_t$ and $\|\hat{\nabla} f_t\|^2$ are estimators rather than exact values, and $\{\nabla f(\theta_t)^\top \mathbf{p}_{t,i}\}_{i=1}^s$ in $\hat{D}_t$ require additional finite-difference approximations. 
Details are provided in Remark A.12 of Appendix A.
Algorithm \ref{alg:pars-opt} is a practical approximation of an idealized version presented in Appendix A. 
Theorem \ref{theorem:convergence} establishes the convergence guarantee for this idealized algorithm, giving an $\mathcal{O}(1/T^2)$ rate under smooth convex assumptions. In comparison, Theorem A.10 shows that Sign-OPT attains an $\mathcal{O}((\ln T)/T)$ rate, indicating that the idealized PARS-OPT converges faster than Sign-OPT.

\begin{theorem}
    \label{theorem:convergence}
    
    Let $\theta^*$ denote the optimal solution of Problem \eqref{eq:goal_OPT}, and let $\theta_0$, $\theta_T$, $\gamma_0$, and $\zeta_t$ denote the corresponding quantities in the idealized version of Algorithm \ref{alg:pars-opt}. Assuming that $f(\cdot)$ is smooth and convex, we have
    {\small
    \begin{align}
        \mathbb{E} \left[ \left( f(\theta_T) - f(\theta^*) \right) \left( 1 + \frac{\sqrt{\gamma_0}}{2} \sum_{t=0}^{T-1} \sqrt{\zeta_t} \right)^2 \right] \leq \notag \\ 
        f(\theta_0) - f(\theta^*) + \frac{\gamma_0}{2} \|\theta_0 - \theta^*\|^2.
        \label{eq:convergence_rate}
    \end{align}
    }
\end{theorem}
The proof is given in Appendix A (Theorem A.11).

\section{Experiments}
\label{sec:Experiments}

\begin{table*}[!t]
	\small
	\tabcolsep=1.3mm
	\centering
	\begin{threeparttable}
		\begin{tabular}{c|lcccccc@{\hspace{2em}}ccccccc}
			\toprule
			\multicolumn{1}{c}{} & \multicolumn{1}{c}{Method} & with  & \multicolumn{5}{c}{Untargeted Attack} & \multicolumn{7}{c}{Targeted Attack} \\
			& & D.R.\tnote{1} & 2K & 4K & 6K & 8K & 10K &  2K & 4K & 6K &8K & 10K & 15K & 20K \\
			\midrule
			\multirow{21}{*}{\rotatebox{90}{Inception-v4}}
			& HSJA & $\times$ & 44.53 & 26.31 & 17.92 & 14.19 & 11.65  & 79.00 & 60.90 & 47.25 & 39.19 & 32.95 & 24.55 & 19.52\\
			& TA & $\times$ & 42.23 & 25.86 & 17.80 & 14.17 & 11.69  & 61.99 & 47.07 & 37.16 & 31.51 & 27.11& 21.08 & 17.32\\
			& Sign-OPT & $\times$ & 48.23 & 23.27 & 14.97 & 11.07 & 8.79  & 65.20 & 48.33 & 38.49 & 32.10 & 27.53 & 20.39 & 16.28\\
			& GeoDA & $\times$ & 20.12 & 14.33 & 12.49 & 11.01 & 9.69  & - & - & - & - & - & - & -\\
			& Evolutionary & $\times$ & 42.66 & 25.32 & 17.60 & 13.38 & 10.84  & 65.06 & 48.37 & 38.72 & 32.12 & 27.39  & 19.94 & 15.61\\
			& SurFree & $\times$ & 38.48 & 26.35 & 20.17 & 16.37 & 13.82  & 74.89 & 61.16 & 51.56 & 44.48 & 39.00  & 29.35 & 23.15\\
			& AHA & $\checkmark$ & 42.06 & 23.52 & 15.41 & 11.10 & 8.52  & 54.12 & 36.09 & 26.46 & 20.50 & 16.49 & 10.86 & 8.12\\
			& QEBA & $\checkmark$ & 16.54 & 8.08 & 5.82 & 4.26 & 3.66  & 58.31 & 37.68 & 28.56 & 21.74 & 18.00 & 12.07 & 9.25\\
			& CGBA-H & $\checkmark$ & 15.12 & 7.83 & 5.86 & 4.61 & 4.10  & 56.32 & 37.82 & 29.69 & 23.86 & 20.00 & 14.31 & 11.56\\
			& SQBA\textsubscript{\tiny IncResV2} & $\times$ & 19.03 & 12.80 & 10.01 & 8.43 & 7.42  & - & - & - & - & -  & - & -\\
			& BBA\textsubscript{\tiny IncResV2} & $\times$ & 28.44 & 20.74 & 17.37 & 15.47 & 14.19  & 56.28 & 44.98 & 38.43 & 34.07 & 30.94  & 25.76 & 22.63\\
			& Prior-Sign-OPT\textsubscript{\tiny IncResV2} & $\times$ & 42.40 & 17.16 & 10.19 & 7.36 & 5.84  & 55.42 & 37.00 & 28.14 & 22.96 & 19.51 & 14.36 & 11.66 \\
			& Prior-Sign-OPT\textsubscript{\tiny IncResV2\&Xception} & $\times$ & 37.10 & 12.57 & 7.10 & 5.19 & 4.20  & 49.37 & 31.34 & 23.67 & 19.32 & 16.70 & 12.82 & 10.77 \\
			& Prior-OPT\textsubscript{\tiny IncResV2} & $\times$ & 18.13 & 6.80 & 5.15 & 4.45 & 4.03  & 49.84 & 36.80 & 31.04 & 27.60 & 25.28 & 21.84 & 19.80\\
			& Prior-OPT\textsubscript{\tiny IncResV2\&Xception} & $\times$ & 13.42 & 4.49 & 3.64 & 3.32 & 3.12  & \B 42.63 & 30.32 & 25.60 & 23.01 & 21.44 & 19.19 & 17.98 \\
			
			\cdashline{2-15}
			\addlinespace[0.1em] 

			& \cellcolor{TableBlue!20}ARS-OPT & \cellcolor{TableBlue!20}$\times$  & \cellcolor{TableBlue!20}46.60 & \cellcolor{TableBlue!20}24.24 & \cellcolor{TableBlue!20}15.74 & \cellcolor{TableBlue!20}11.68 & \cellcolor{TableBlue!20}9.30  & \cellcolor{TableBlue!20}65.53 & \cellcolor{TableBlue!20}46.60 & \cellcolor{TableBlue!20}35.84 & \cellcolor{TableBlue!20}28.84 & \cellcolor{TableBlue!20}24.02 & \cellcolor{TableBlue!20}16.63 & \cellcolor{TableBlue!20}12.69\\
			& \cellcolor{TableBlue!20}PARS-OPT\textsubscript{\tiny IncResV2} & \cellcolor{TableBlue!20}$\times$ & \cellcolor{TableBlue!20}14.02 & \cellcolor{TableBlue!20}6.31 & \cellcolor{TableBlue!20}4.93 & \cellcolor{TableBlue!20}4.24 & \cellcolor{TableBlue!20}3.82  & \cellcolor{TableBlue!20}49.37 & \cellcolor{TableBlue!20}33.88 & \cellcolor{TableBlue!20}26.91 & \cellcolor{TableBlue!20}22.72 & \cellcolor{TableBlue!20}19.94 & \cellcolor{TableBlue!20}16.06 & \cellcolor{TableBlue!20}13.67\\
			& \cellcolor{TableBlue!20}PARS-OPT\textsubscript{\tiny IncResV2\&Xception} & \cellcolor{TableBlue!20}$\times$ & \B \cellcolor{TableBlue!20}9.91 & \B \cellcolor{TableBlue!20}4.41 & \B \cellcolor{TableBlue!20}3.62 & \B \cellcolor{TableBlue!20}3.28 & \B \cellcolor{TableBlue!20}3.05  & \cellcolor{TableBlue!20}43.91 & \B \cellcolor{TableBlue!20}28.16 & \B \cellcolor{TableBlue!20}22.56 & \cellcolor{TableBlue!20}19.36 & \cellcolor{TableBlue!20}17.23 & \cellcolor{TableBlue!20}14.13 & \cellcolor{TableBlue!20}12.32 \\
			& \cellcolor{TableBlue!20}ARS-OPT-S & \cellcolor{TableBlue!20}$\checkmark$  & \cellcolor{TableBlue!20}25.02 & \cellcolor{TableBlue!20}10.38 & \cellcolor{TableBlue!20}6.46 & \cellcolor{TableBlue!20}4.85 & \cellcolor{TableBlue!20}3.92  & \cellcolor{TableBlue!20}59.15 & \cellcolor{TableBlue!20}37.52 & \cellcolor{TableBlue!20}26.37 & \cellcolor{TableBlue!20}19.62 & \cellcolor{TableBlue!20}15.37 & \B \cellcolor{TableBlue!20}9.94 & \B \cellcolor{TableBlue!20}7.38\\
			& \cellcolor{TableBlue!20}PARS-OPT-S\textsubscript{\tiny IncResV2} & \cellcolor{TableBlue!20}$\checkmark$ & \cellcolor{TableBlue!20}19.55 & \cellcolor{TableBlue!20}7.82 & \cellcolor{TableBlue!20}5.36 & \cellcolor{TableBlue!20}4.23 & \cellcolor{TableBlue!20}3.54  & \cellcolor{TableBlue!20}55.18 & \cellcolor{TableBlue!20}34.12 & \cellcolor{TableBlue!20}24.22 & \cellcolor{TableBlue!20}18.73 & \B \cellcolor{TableBlue!20}15.02 & \cellcolor{TableBlue!20}10.18 & \cellcolor{TableBlue!20}7.84\\
			& \cellcolor{TableBlue!20}PARS-OPT-S\textsubscript{\tiny IncResV2\&Xception} & \cellcolor{TableBlue!20}$\checkmark$ & \cellcolor{TableBlue!20}20.52 & \cellcolor{TableBlue!20}7.25 & \cellcolor{TableBlue!20}5.02 & \cellcolor{TableBlue!20}4.05 & \cellcolor{TableBlue!20}3.45  & \cellcolor{TableBlue!20}55.28 & \cellcolor{TableBlue!20}33.30 & \cellcolor{TableBlue!20}23.70 & \B \cellcolor{TableBlue!20}18.64 & \cellcolor{TableBlue!20}15.31 & \cellcolor{TableBlue!20}10.78 & \cellcolor{TableBlue!20}8.33 \\
			
			\midrule
			\multirow{21}{*}{\rotatebox{90}{Swin Transformer}} 
			& HSJA & $\times$ & 45.86 & 27.32 & 17.92 & 13.50 & 10.64  & 50.96 & 39.26 & 30.66 & 25.64 & 21.73 & 16.19 & 12.75\\
			& TA & $\times$ & 46.73 & 27.85 & 18.02 & 13.38 & 10.51  & 40.72 & 31.92 & 25.88 & 22.25 & 19.45& 15.52 & 12.89\\
			& Sign-OPT & $\times$ & 53.40 & 26.41 & 16.93 & 12.41 & 9.90  & 44.91 & 35.98 & 30.89 & 27.52 & 25.27 & 21.84 & 19.95\\
			& GeoDA & $\times$ & 36.92 & 28.03 & 24.54 & 21.59 & 19.12  & - & - & - & - & - & - & -\\
			& Evolutionary & $\times$ & 49.24 & 31.19 & 23.04 & 18.60 & 15.74  & 51.71 & 38.29 & 31.23 & 26.85 & 23.76  & 19.28 & 16.56\\
			& SurFree & $\times$  & 34.28 & 23.58 & 18.37 & 15.18 & 13.06  & 61.31 & 47.67 & 39.39 & 33.84 & 29.73  & 22.96 & 18.73\\
			& AHA & $\checkmark$ & 46.76 & 30.37 & 23.35 & 19.39 & 17.02  & 36.11 & 28.04 & 23.68 & 20.78 & 18.76 & 15.51 & 13.72\\
			& QEBA & $\checkmark$ & 31.11 & 16.99 & 12.07 & 8.46 & 7.02  & 42.99 & 30.31 & 24.38 & 19.40 & 16.52 & 11.58 & 8.91\\
			& CGBA-H & $\checkmark$ & 29.24 & 17.01 & 12.60 & 9.26 & 7.81  & 37.81 & 27.83 & 23.19 & 19.67 & 17.17 & 13.10 & 10.83\\
			& SQBA\textsubscript{\tiny ResNet50} & $\times$ & 20.40 & 13.40 & 10.33 & 8.62 & 7.56  & - & - & - & - & -  & - & -\\
			& BBA\textsubscript{\tiny ResNet50} & $\times$  & 29.37 & 20.94 & 17.59 & 15.47 & 14.08  & 35.28 & 28.45 & 24.65 & 22.16 & 20.34  & 17.54 & 15.98\\
			& Prior-Sign-OPT\textsubscript{\tiny ResNet50} & $\times$ & 52.88 & 26.19 & 16.45 & 11.88 & 9.25  & 43.88 & 34.32 & 29.23 & 26.06 & 23.86 & 20.52 & 18.66 \\
			& Prior-Sign-OPT\textsubscript{\tiny ResNet50\&ConViT} & $\times$ & 43.06 & 17.96 & 10.91 & 7.90 & 6.33  & 43.20 & 33.48 & 28.21 & 24.99 & 22.84 & 19.66 & 17.94 \\
			& Prior-OPT\textsubscript{\tiny ResNet50} & $\times$ & 39.45 & 20.26 & 14.13 & 11.24 & 9.62  & 42.96 & 33.51 & 28.64 & 25.67 & 23.72 & 20.86 & 19.45\\
			& Prior-OPT\textsubscript{\tiny ResNet50\&ConViT} & $\times$ & 17.98 & 8.66 & 6.45 & 5.45 & 4.90  & 39.62 & 30.27 & 25.75 & 23.12 & 21.45 & 19.27 & 18.33 \\
			
			\cdashline{2-15}\addlinespace[0.1em] 
			& \cellcolor{TableBlue!20}ARS-OPT & \cellcolor{TableBlue!20}$\times$  & \cellcolor{TableBlue!20}41.91 & \cellcolor{TableBlue!20}20.04 & \cellcolor{TableBlue!20}12.76 & \cellcolor{TableBlue!20}9.26 & \cellcolor{TableBlue!20}7.21  & \cellcolor{TableBlue!20}38.85 & \cellcolor{TableBlue!20}26.14 & \cellcolor{TableBlue!20}19.70 & \cellcolor{TableBlue!20}15.67 & \cellcolor{TableBlue!20}12.99 & \cellcolor{TableBlue!20}9.10 & \cellcolor{TableBlue!20}6.96\\
			& \cellcolor{TableBlue!20}PARS-OPT\textsubscript{\tiny ResNet50} & \cellcolor{TableBlue!20}$\times$ & \cellcolor{TableBlue!20}29.26 & \cellcolor{TableBlue!20}12.77 & \cellcolor{TableBlue!20}8.41 & \cellcolor{TableBlue!20}6.22 & \cellcolor{TableBlue!20}5.01  & \cellcolor{TableBlue!20}38.01 & \cellcolor{TableBlue!20}25.72 & \cellcolor{TableBlue!20}19.72 & \cellcolor{TableBlue!20}15.73 & \cellcolor{TableBlue!20}13.15 & \cellcolor{TableBlue!20}9.39 & \cellcolor{TableBlue!20}7.29\\
			& \cellcolor{TableBlue!20}PARS-OPT\textsubscript{\tiny ResNet50\&ConViT} & \cellcolor{TableBlue!20}$\times$ & \B \cellcolor{TableBlue!20}12.73 & \B \cellcolor{TableBlue!20}6.11 & \B \cellcolor{TableBlue!20}4.56 & \B \cellcolor{TableBlue!20}3.73 & \B \cellcolor{TableBlue!20}3.23  & \cellcolor{TableBlue!20}36.53 & \cellcolor{TableBlue!20}23.60 & \cellcolor{TableBlue!20}17.98 & \cellcolor{TableBlue!20}14.50 & \cellcolor{TableBlue!20}12.20 & \cellcolor{TableBlue!20}8.61 & \cellcolor{TableBlue!20}6.90 \\
			& \cellcolor{TableBlue!20}ARS-OPT-S & \cellcolor{TableBlue!20}$\checkmark$  & \cellcolor{TableBlue!20}23.04 & \cellcolor{TableBlue!20}10.61 & \cellcolor{TableBlue!20}6.88 & \cellcolor{TableBlue!20}5.06 & \cellcolor{TableBlue!20}3.99  & \B \cellcolor{TableBlue!20}34.77 & \B \cellcolor{TableBlue!20}20.92 & \B \cellcolor{TableBlue!20}14.46 & \B \cellcolor{TableBlue!20}10.71 & \B \cellcolor{TableBlue!20}8.31 & \B \cellcolor{TableBlue!20}5.24 & \B \cellcolor{TableBlue!20}3.79\\
			& \cellcolor{TableBlue!20}PARS-OPT-S\textsubscript{\tiny ResNet50} & \cellcolor{TableBlue!20}$\checkmark$ & \cellcolor{TableBlue!20}23.91 & \cellcolor{TableBlue!20}10.96 & \cellcolor{TableBlue!20}7.23 & \cellcolor{TableBlue!20}5.40 & \cellcolor{TableBlue!20}4.31  & \cellcolor{TableBlue!20}37.10 & \cellcolor{TableBlue!20}22.92 & \cellcolor{TableBlue!20}15.85 & \cellcolor{TableBlue!20}11.88 & \cellcolor{TableBlue!20}9.42 & \cellcolor{TableBlue!20}6.03 & \cellcolor{TableBlue!20}4.30\\
			& \cellcolor{TableBlue!20}PARS-OPT-S\textsubscript{\tiny ResNet50\&ConViT} & \cellcolor{TableBlue!20}$\checkmark$ & \cellcolor{TableBlue!20}19.84 & \cellcolor{TableBlue!20}8.74 & \cellcolor{TableBlue!20}6.09 & \cellcolor{TableBlue!20}4.68 & \cellcolor{TableBlue!20}3.85  & \cellcolor{TableBlue!20}37.06 & \cellcolor{TableBlue!20}22.56 & \cellcolor{TableBlue!20}16.26 & \cellcolor{TableBlue!20}12.26 & \cellcolor{TableBlue!20}9.83 & \cellcolor{TableBlue!20}6.54 & \cellcolor{TableBlue!20}4.80 \\
			
			\bottomrule
		\end{tabular}
		\begin{tablenotes}
			\footnotesize
			\item[1] D.R. denotes the use of dimension reduction technique.
		\end{tablenotes}
	\end{threeparttable}
	\caption{Mean $\ell_2$ distortions of different query budgets on the ImageNet dataset.}
	\label{tab:ImageNet_normal_models_result}
\end{table*}

\begin{table*}[!t]
	\small
	\tabcolsep=1.3mm
	\centering   
	
	\begin{tabular}{lccccc@{\hspace{2em}}ccccc}
		\toprule
		\multicolumn{1}{c}{Method}  & \multicolumn{5}{c}{Mean $\ell_2$ Distortions} & \multicolumn{5}{c}{Attack Success Rate} \\
		& 2K & 4K & 6K & 8K & 10K &  2K & 4K & 6K & 8K & 10K \\
		\midrule
		Sign-OPT & 49.44 & 42.29 & 38.93 & 37.02 & 35.71  & 15.2\% & 16.3\% & 18.0\% & 19.1\% & 19.4\%\\
		\cdashline{1-11} 
		\addlinespace[0.1em] 
		Prior-OPT\textsubscript{\tiny ResNet50}& 27.38 & 21.52 & 19.34 & 18.52 & 18.15  & 34.9\% & 44.9\% & 48.7\% & 50.6\% & 51.4\% \\
		Prior-OPT\textsubscript{\tiny ConViT} & 21.27 & 16.54 & 15.14 & 14.62 & 14.36  & 43.3\% & 54.7\% & 57.5\% & 58.8\% & 58.9\% \\
		Prior-OPT\textsubscript{\tiny ResNet50\&ConViT} & \B 18.09 & 12.66 & 11.22 & 10.72 & 10.43  & \B 50.1\% & \B 65.6\% & 70.2\% & 72.2\% & 73.4\%  \\
		
		\cdashline{1-11} 
		\addlinespace[0.1em] 
		\rowcolor{gray!25} ARS-OPT  & 47.42 & 37.08 & 31.00 & 26.97 & 24.02  & 16.2\% & 20.2\% & 24.7\% & 28.2\% & 30.7\% \\
		
		\cdashline{1-11} 
		\addlinespace[0.1em] 
		\rowcolor{TableBlue!20} PARS-OPT\textsubscript{\tiny ResNet50} & 26.18 & 17.65 & 14.45 & 12.69 & 11.55  & 36.2\% & 49.8\% & 56.8\% & 62.2\% & 65.8\% \\
		\rowcolor{TableBlue!20} PARS-OPT\textsubscript{\tiny ConViT} & 20.80 & 14.89 & 12.68 & 11.47 & 10.61  & 43.3\% & 57.0\% & 62.0\% & 66.5\% & 69.1\% \\
		\rowcolor{TableBlue!20} PARS-OPT\textsubscript{\tiny ResNet50\&ConViT} & 18.67 & \B 12.25 & \B 10.16 & \B 9.07 & \B 8.38  & 48.7\% & 65.1\% & \B 72.6\% & \B 76.0\% & \B 78.8\%  \\
		\bottomrule
	\end{tabular}
	\caption{The experimental results of attacking against CLIP with the backbone of ViT-L/14.}
	\label{tab:CLIP}
\end{table*}

\begin{figure*}[t]
    \centering
    \begin{subfigure}{0.46\textwidth}
        \includegraphics[width=\linewidth]{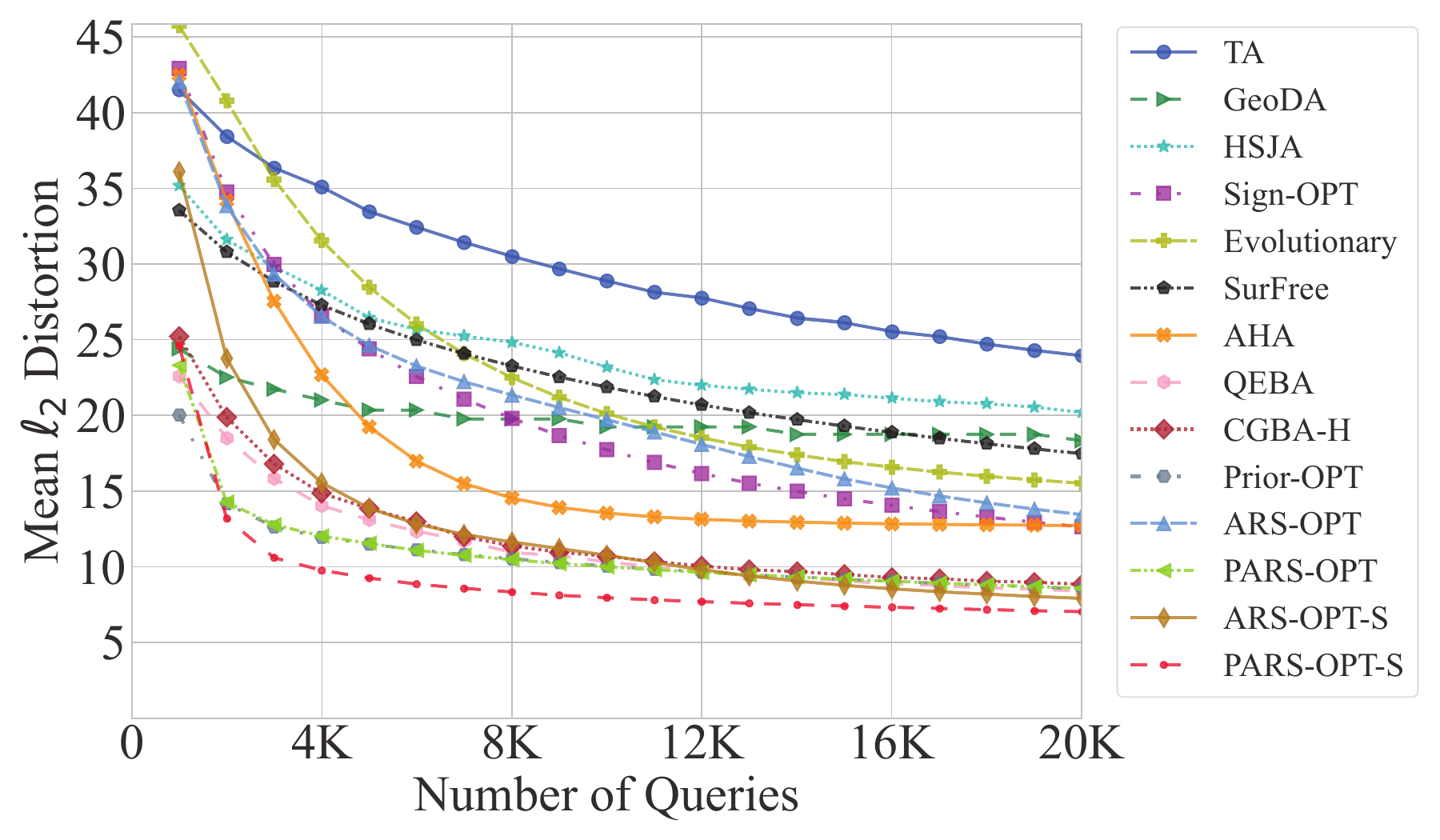}
        \caption{AT$_{\text{ResNet50}, \epsilon_\infty = 8/255}$(ImageNet)}
    \end{subfigure}
    \begin{subfigure}{0.46\textwidth}
        \includegraphics[width=\linewidth]{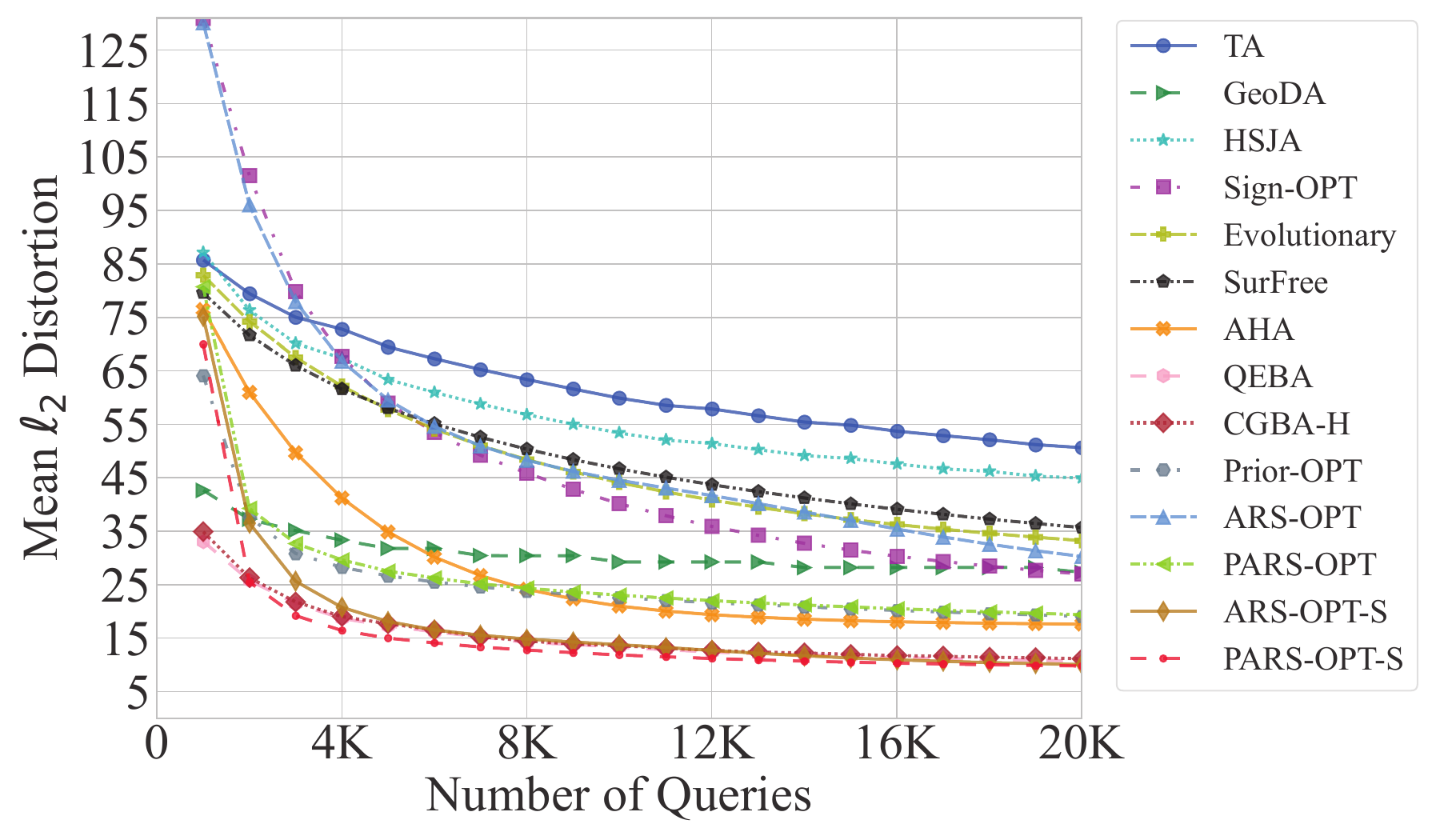}
        \caption{MIMIR$_{\text{ViT}, \epsilon_\infty = 4/255}$(ImageNet)}
    \end{subfigure}
    \begin{subfigure}{0.46\textwidth}
        \includegraphics[width=\linewidth]{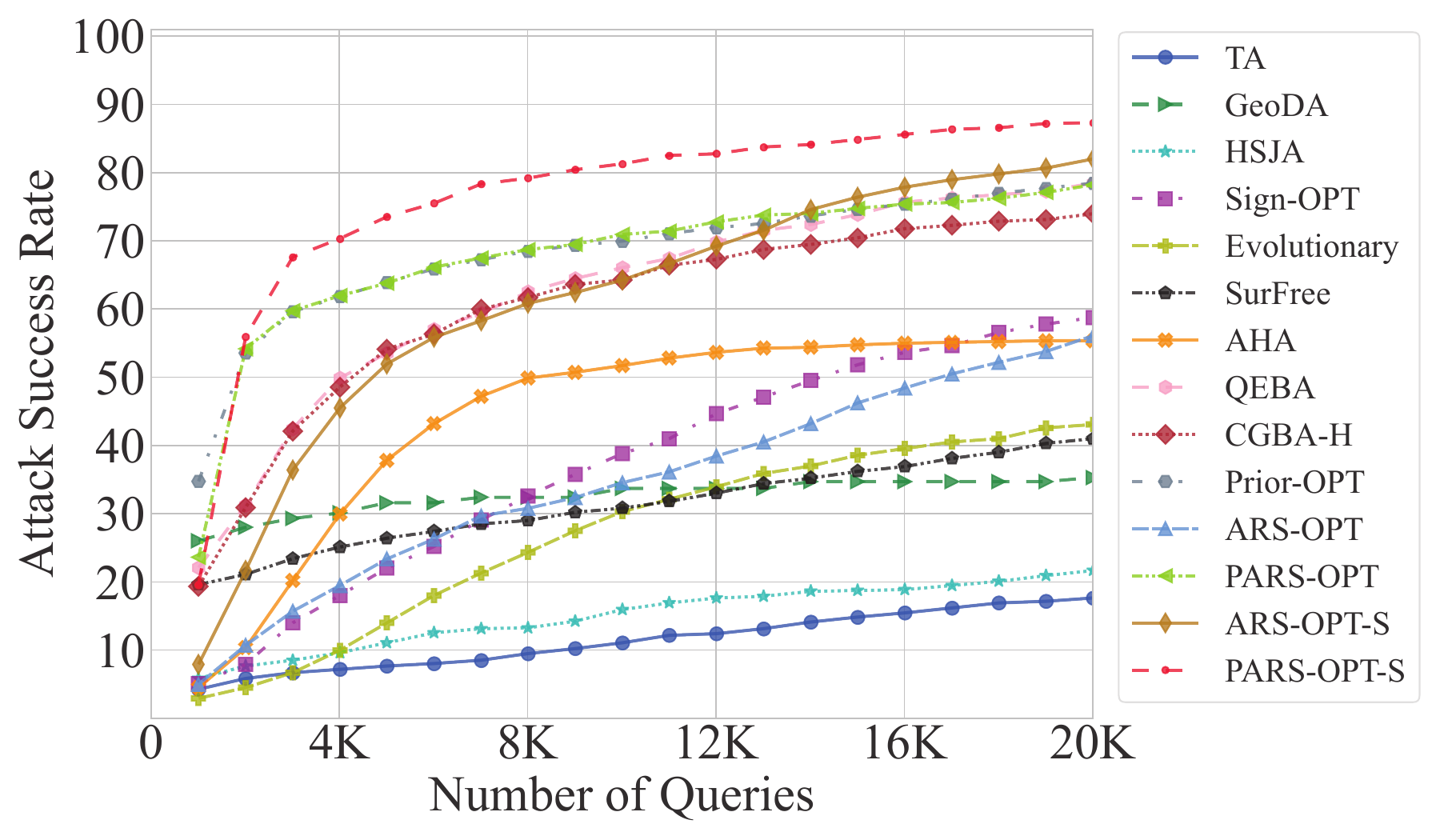}
        \caption{AT$_{\text{ResNet50}, \epsilon_\infty = 8/255}$(ImageNet)}
    \end{subfigure}
    \begin{subfigure}{0.46\textwidth}
        \includegraphics[width=\linewidth]{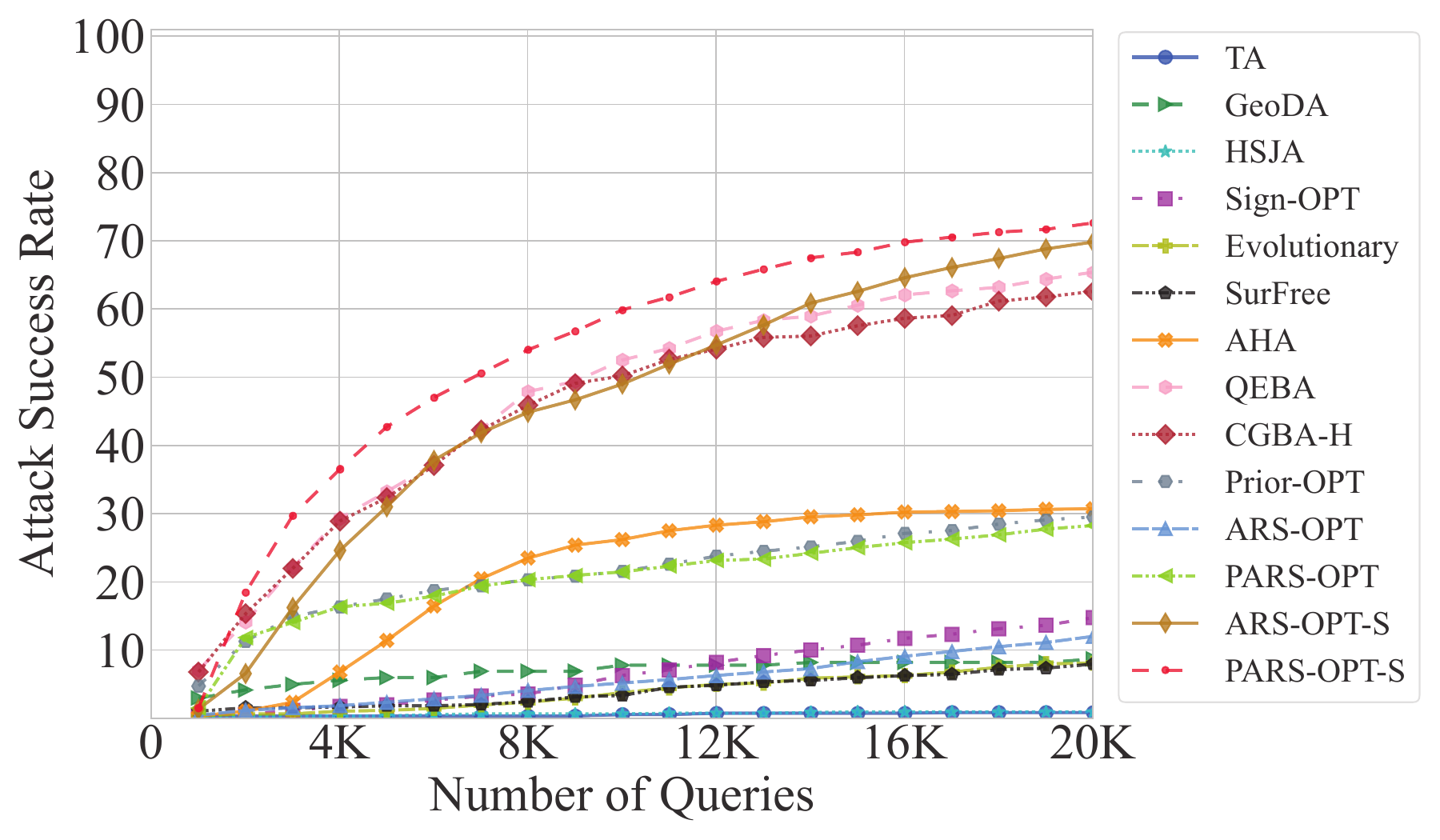}
        \caption{MIMIR$_{\text{ViT}, \epsilon_\infty = 4/255}$(ImageNet)}
    \end{subfigure}
    \caption{Mean distortions and attack success rates of untargeted attacks with $\ell_2$ norm constraint against defense models. The surrogate model of PARS-OPT and Prior-OPT is the adversarially trained ResNet-50 model (PGD, $\epsilon_{\ell_\infty}=4/255$).}
    \label{fig:defense_untargeted_attack}
\end{figure*}

\begin{figure}[ht]
    \centering
    \begin{subfigure}{0.23\textwidth}

        \includegraphics[width=\linewidth]{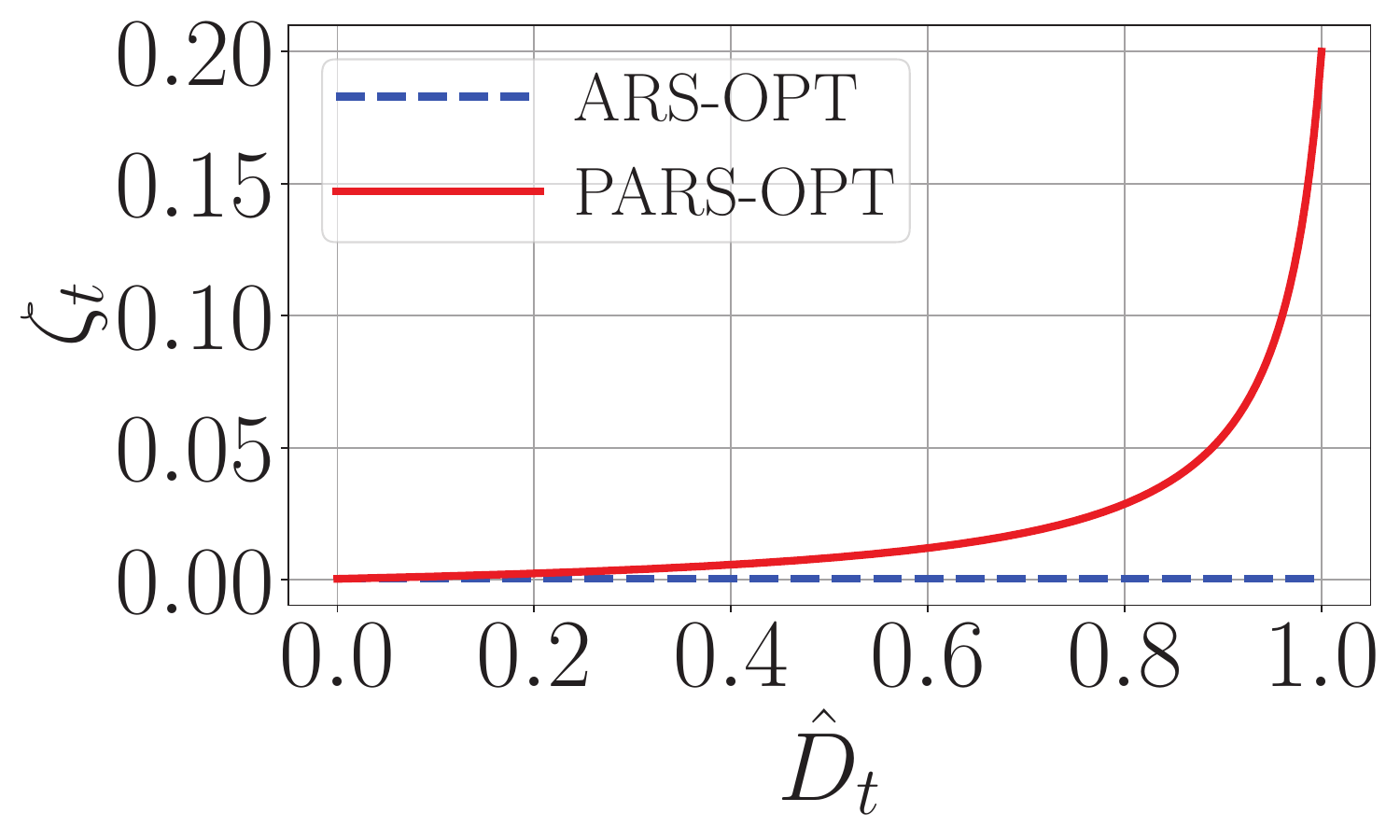}
        \caption{Effect of prior's quality $\hat{D}_t$}
        \label{subfig:D_t_vs_zeta_t}
    \end{subfigure}
    \begin{subfigure}{0.23\textwidth}

        \includegraphics[width=\linewidth]{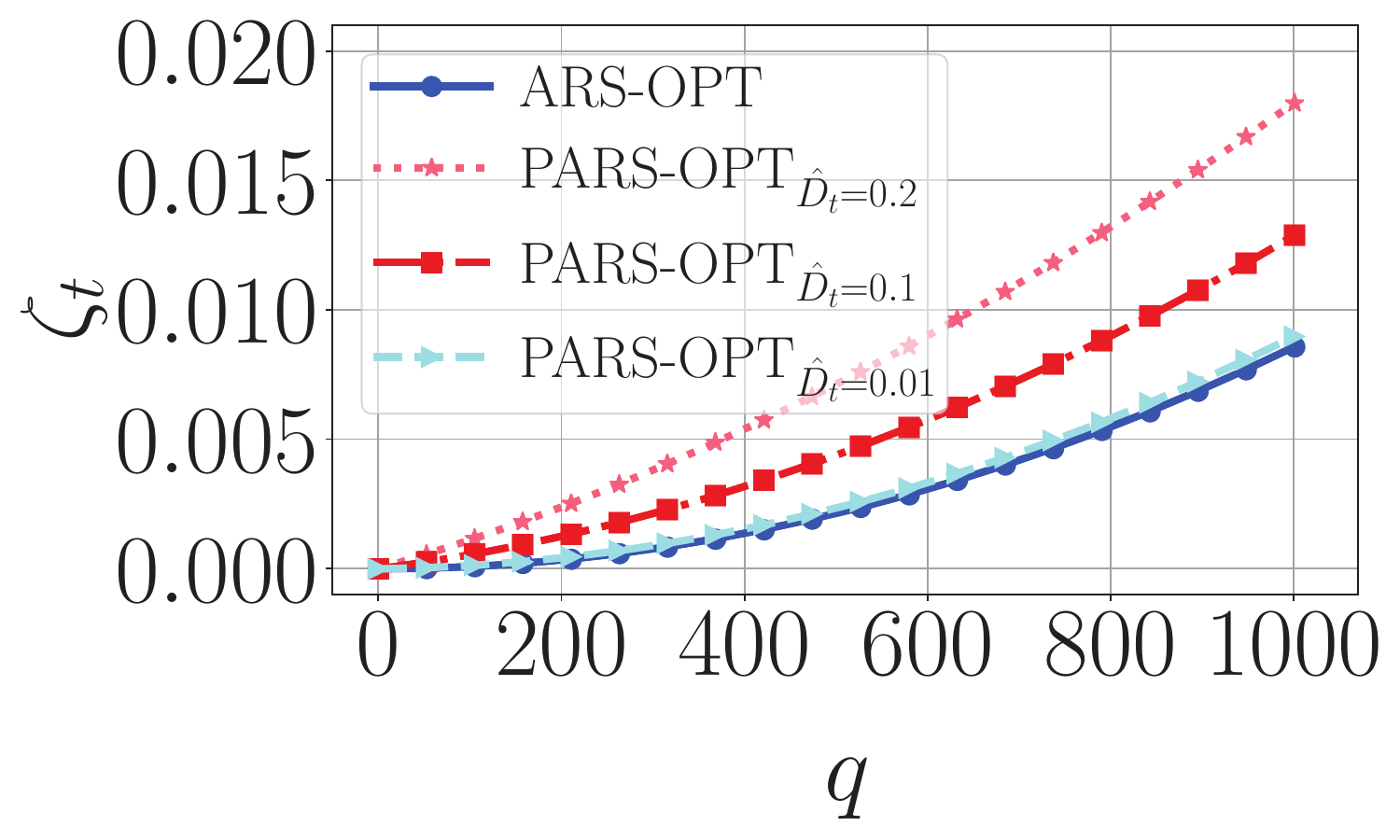}
        \caption{Effect of number of vectors $q$}
        \label{subfig:q_vs_zeta_t}
    \end{subfigure}
    \begin{subfigure}{0.23\textwidth}

        \includegraphics[width=\linewidth]{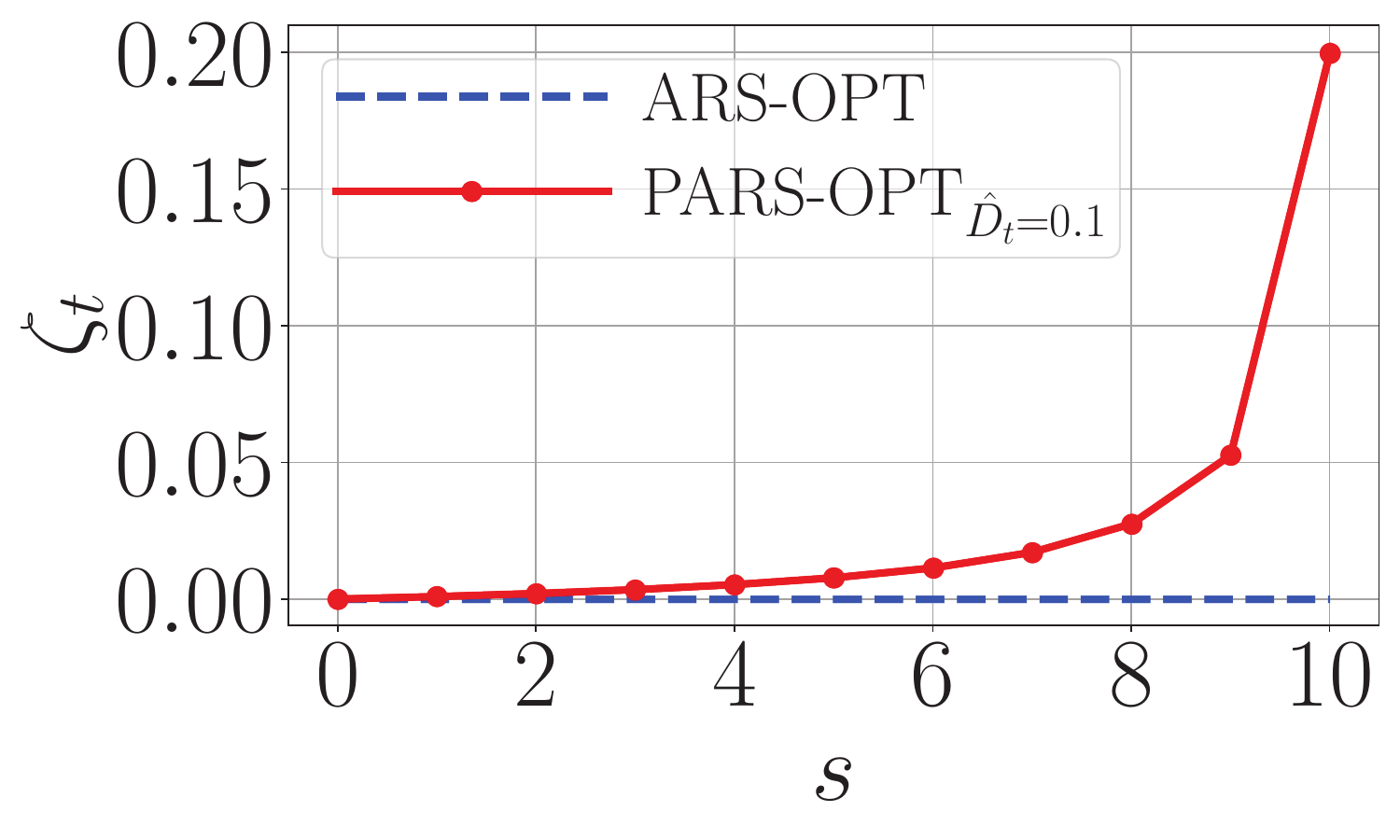}
        \caption{Effect of number of priors $s$}
        \label{subfig:s_vs_zeta_t}
    \end{subfigure}
    \begin{subfigure}{0.23\textwidth}

        \includegraphics[width=\linewidth]{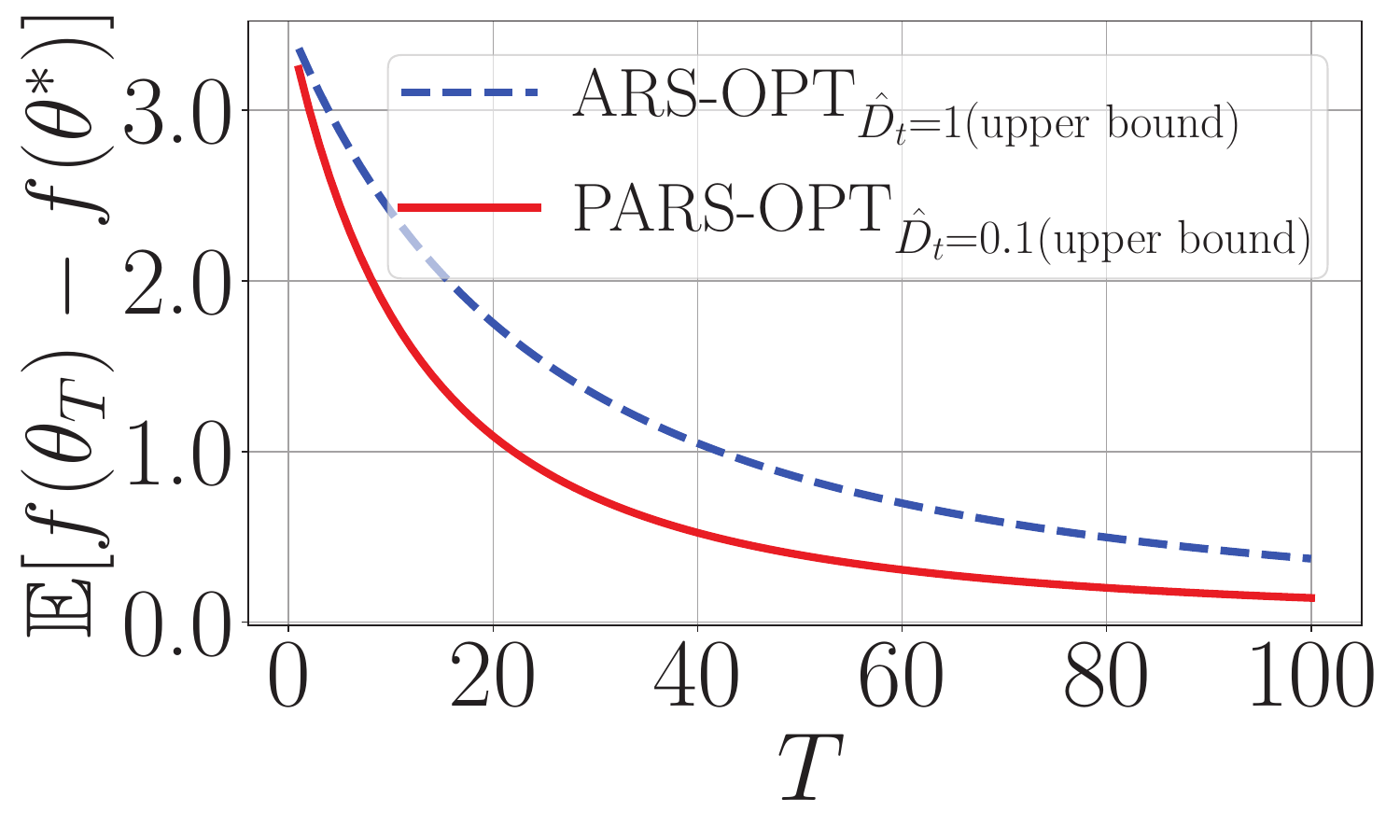}
        \caption{Convergence of (P)ARS-OPT}
        \label{subfig:convergence}
    \end{subfigure}
    \caption{Experimental results of ablation studies.}
\end{figure}
\subsection{Experimental Setting}
\textbf{Dataset.} We evaluate the proposed method on two publicly available datasets, CIFAR-10 \cite{krizhevsky2009learning} and ImageNet \cite{deng2009imagenet}, with images resized to $3 \times 32 \times 32$ and $3 \times 299 \times 299$, respectively. For all experiments, \nn{1000} images are randomly selected from each dataset as test samples for evaluation. In the case of targeted attacks, the target class is defined as $y_\text{adv} = (y + 1) \mod C$, where $y$ denotes the true class. For the same target class, we use the same image $\tilde{\mathbf{x}}$ as the initialization for all methods.

\noindent\textbf{Models.} On the ImageNet dataset, we evaluate two target models: Inception-v4 \cite{szegedy2017inceptionv4} and Swin Transformer \cite{liu2021swin}. 
For Inception-v4 (input resolution $299 \times 299$), we use Inception-ResNet-v2 (IncResV2) and Xception as surrogate models. For Swin Transformer (inputs resized to $224 \times 224$), the surrogate models are ResNet-50 and ConViT \cite{d2021convit}. See Appendix for details.

\noindent\textbf{Baseline Methods.} We compare ARS-OPT and PARS-OPT against baselines, including HSJA, TA, GeoDA, Evolutionary, SurFree, AHA, QEBA, CGBA-H, SQBA, BBA, Sign-OPT, Prior-Sign-OPT and Prior-OPT. In our methods, the suffix ``-S'' (e.g., ARS-OPT-S) means the random vectors $\bu_1,\dots,\bu_{q-s}$ for gradient estimation are drawn from a $3\times56\times56$-dimensional subspace. AHA, QEBA, and CGBA-H also adopt subspace sampling, while SQBA, BBA, Prior-Sign-OPT, Prior-OPT, and PARS-OPT leverage surrogate models, denoted by subscripts; e.g., PARS-OPT\textsubscript{IncResV2} uses Inception-ResNet-v2 as the surrogate model.


\noindent\textbf{Metrics.} We report the mean $\ell_2$ distortion as $\frac{1}{| \mathbf{X} |}\sum_{\mathbf{x} \in \mathbf{X}} \| \mathbf{x}_\text{adv} - \mathbf{x} \|_2$, where $\mathbf{X}$ denotes the test dataset. Additionally, we present the attack success rate (ASR), defined as the proportion of samples with distortions below a threshold $\sqrt{0.001 \times d}$ for a given query budget.

\subsection{Experimental Results on the ImageNet Dataset}

\textbf{Results of Attacks against Undefended Models.} Tables \ref{tab:ImageNet_normal_models_result} and \ref{tab:CLIP} report the results of attacks against undefended models on \nn{1000} ImageNet images. In summary:

(1) In Table \ref{tab:ImageNet_normal_models_result}, PARS-OPT performs the best in untargeted attacks, while ARS-OPT-S achieves state-of-the-art performance in targeted attacks due to its stabilized optimization via the lookahead direction, reducing the risk of local minima.

(2) Table \ref{tab:CLIP} reports untargeted attack results on CLIP (ViT-L/14). Our methods outperform the baselines (Sign-OPT and Prior-OPT) in mean $\ell_2$ distortion and attack success rate.

\noindent\textbf{Results of Attacks against Defense Models.}
We evaluate untargeted attacks against two types of defense models, i.e., adversarial training (AT) \cite{madry2018towards} and MIMIR \cite{xu2023mimir}. MIMIR achieves state-of-the-art performance on RobustBench \cite{croce2021robustbench}. Fig. \ref{fig:defense_untargeted_attack} shows that our methods achieve the best performance on ImageNet.

\subsection{Comprehensive Understanding of (P)ARS-OPT}
In our ablation studies, we perform controlled experiments designed according to our theoretical analysis, using images of dimension $d=3072$. 
Fig. \ref{subfig:D_t_vs_zeta_t} shows the relationship between $\hat{D}_t$ and $\zeta_t$. As $\hat{D}_t$ increases, $\zeta_t$ increases accordingly, which in turn improves the convergence rate of PARS-OPT (Eq.~\eqref{eq:convergence_rate}).
Fig. \ref{subfig:q_vs_zeta_t} illustrates that increasing the number of vectors $q$ used for gradient estimation leads to larger $\zeta_t$ and improved performance.
Fig. \ref{subfig:s_vs_zeta_t} shows that when the priors have equal quality (i.e., identical $\hat{D}_t$ values), increasing their number leads to larger $\zeta_t$, thereby improving attack efficiency.
Fig. \ref{subfig:convergence} shows that when the prior is effective, even with a small $\hat{D}_t$, PARS-OPT achieves a lower convergence bound than ARS-OPT, indicating better potential performance.

\section{Conclusion}
We propose a novel hard-label attack approach, comprising two algorithms---ARS-OPT and PARS-OPT---that accelerate convergence and improve attack success rates by leveraging a lookahead direction and transfer-based priors. We provide convergence guarantees through theoretical analysis and validate our methods with extensive experiments, demonstrating improvements over 13 state-of-the-art approaches.

\section*{Acknowledgments}
This work was supported by Zhejiang Provincial Natural Science Foundation of China under Grant No. LMS25F020005, and by the Key R\&D Program of Zhejiang Province under Grant No. 2024C01164.

\bibliography{aaai2026}

@INPROCEEDINGS{rahmati2020geoda,
  author={Rahmati, Ali and Moosavi-Dezfooli, Seyed-Mohsen and Frossard, Pascal and Dai, Huaiyu},
  booktitle={IEEE/CVF Conference on Computer Vision and Pattern Recognition}, 
  title={{GeoDA}: A Geometric Framework for Black-Box Adversarial Attacks}, 
  year={2020},
  pages={8443-8452},
  doi={10.1109/CVPR42600.2020.00847}
}

@inproceedings{ma2021simulator,
	author={Ma, Chen and Chen, Li and Yong, Jun-Hai},
	booktitle={IEEE/CVF Conference on Computer Vision and Pattern Recognition}, 
	title={Simulating Unknown Target Models for Query-Efficient Black-box Attacks}, 
	year={2021},
	pages={11830-11839},
	doi={10.1109/CVPR46437.2021.01166}
}

@inproceedings{cheng2021ontheconvergence,
 author = {Cheng, Shuyu and Wu, Guoqiang and Zhu, Jun},
 booktitle = {Advances in Neural Information Processing Systems},
 //editor = {M. Ranzato and A. Beygelzimer and Y. Dauphin and P.S. Liang and J. Wortman Vaughan},
 pages = {14620--14631},
 publisher = {Curran Associates, Inc.},
 title = {On the Convergence of Prior-Guided Zeroth-Order Optimization Algorithms},
 url = {https://proceedings.neurips.cc/paper_files/paper/2021/file/7aaece81f2d731fbf8ee0ad3521002ac-Paper.pdf},
 volume = {34},
 year = {2021}
}

@inproceedings{cheng2020sign,
title={{Sign-OPT}: A Query-Efficient Hard-label Adversarial Attack},
author={Minhao Cheng and Simranjit Singh and Patrick H. Chen and Pin-Yu Chen and Sijia Liu and Cho-Jui Hsieh},
booktitle={International Conference on Learning Representations},
year={2020},
url={https://openreview.net/forum?id=SklTQCNtvS}
}

@inproceedings {brunner2019guessing,
author = {Brunner, Thomas and Diehl, Frederik and Le, Michael Truong and Knoll, Alois},
booktitle = {IEEE/CVF International Conference on Computer Vision},
title = {Guessing Smart: Biased Sampling for Efficient Black-Box Adversarial Attacks},
year = {2019},
pages = {4957-4965},
doi = {10.1109/ICCV.2019.00506},
//url = {https://doi.ieeecomputersociety.org/10.1109/ICCV.2019.00506},
publisher = {IEEE Computer Society},
//address = {Los Alamitos, CA, USA},
//month =Nov
}

@inproceedings{park2024sqba,
  author={Park, Jeonghwan and Miller, Paul and McLaughlin, Niall},
  booktitle={IEEE/CVF Winter Conference on Applications of Computer Vision}, 
  title={Hard-label based Small Query Black-box Adversarial Attack}, 
  year={2024},
  pages={3974-3983},
  doi={10.1109/WACV57701.2024.00394}}

@inproceedings{deng2009imagenet,
  author={Deng, Jia and Dong, Wei and Socher, Richard and Li, Li-Jia and Kai Li and Li Fei-Fei},
  booktitle={IEEE Conference on Computer Vision and Pattern Recognition}, 
  title={{ImageNet}: A Large-Scale Hierarchical Image Database}, 
  year={2009},
  pages={248-255},
  doi={10.1109/CVPR.2009.5206848}}

@Techreport{krizhevsky2009learning,
 title = {Learning Multiple Layers of Features from Tiny Images},
 author = {Krizhevsky, Alex and Hinton, Geoffrey},
 //address = {Toronto, Ontario},
 institution = {University of Toronto},
 number = {0},
 publisher = {Technical report, University of Toronto},
 year = {2009},
 url = {https://www.cs.toronto.edu/~kriz/learning-features-2009-TR.pdf}
}

@inproceedings{brendel2018decisionbased,
title={Decision-Based Adversarial Attacks: Reliable Attacks Against Black-Box Machine Learning Models},
author={Wieland Brendel and Jonas Rauber and Matthias Bethge},
booktitle={International Conference on Learning Representations},
year={2018},
url={https://openreview.net/forum?id=SyZI0GWCZ},
}

@inproceedings{madry2018towards,
title={Towards Deep Learning Models Resistant to Adversarial Attacks},
author={Aleksander Madry and Aleksandar Makelov and Ludwig Schmidt and Dimitris Tsipras and Adrian Vladu},
booktitle={International Conference on Learning Representations},
year={2018},
url={https://openreview.net/forum?id=rJzIBfZAb},
}

@inproceedings{zhang2024qedba,
  author={Zhang, Zhuosheng and Ahmed, Noor and Yu, Shucheng},
  booktitle={International Conference on Computing, Networking and Communications}, 
  title={{QE-DBA}: Query-Efficient Decision-Based Adversarial Attacks via Bayesian Optimization}, 
  year={2024},
  pages={783--788},
  doi={10.1109/ICNC59896.2024.10555954}}

@inproceedings{ma2025boosting,
title={Boosting Ray Search Procedure of Hard-label Attacks with Transfer-based Priors},
author={Chen Ma and Xinjie Xu and Shuyu Cheng and Qi Xuan},
booktitle={International Conference on Learning Representations},
year={2025},
url={https://openreview.net/forum?id=tIBAOcAvn4}
}

@inproceedings{goodfellow6572explaining,
	author    = {Ian J. Goodfellow and Jonathon Shlens and Christian Szegedy},
	title     = {Explaining and Harnessing Adversarial Examples},
	booktitle = {International Conference on Learning Representations},
	year      = {2015},
	url       = {http://arxiv.org/abs/1412.6572},
}

@inproceedings{reza2023cgba,
  author={Reza, Md Farhamdur and Rahmati, Ali and Wu, Tianfu and Dai, Huaiyu},
  booktitle={IEEE/CVF International Conference on Computer Vision}, 
  title={{CGBA}: Curvature-aware Geometric Black-box Attack}, 
  year={2023},
  pages={124-133},
  doi={10.1109/ICCV51070.2023.00018}}

@InProceedings{d2021convit,
	title={{ConViT}: Improving Vision Transformers with Soft Convolutional Inductive Biases},
	author={D'Ascoli, St{\'e}phane and Touvron, Hugo and Leavitt, Matthew L and Morcos, Ari S and Biroli, Giulio and Sagun, Levent},
	booktitle={International Conference on Machine Learning},
	pages= {2286--2296},
	year={2021},
	//editor={Meila, Marina and Zhang, Tong},
	volume ={139},
	series={Proceedings of Machine Learning Research},
	//month ={18--24 Jul},
	publisher={PMLR},
	//pdf={http://proceedings.mlr.press/v139/d-ascoli21a/d-ascoli21a.pdf},
	//url={https://proceedings.mlr.press/v139/d-ascoli21a.html},
}

@inproceedings {liu2021swin,
author = {Liu, Ze and Lin, Yutong and Cao, Yue and Hu, Han and Wei, Yixuan and Zhang, Zheng and Lin, Stephen and Guo, Baining},
booktitle = {IEEE/CVF International Conference on Computer Vision},
title = {{Swin Transformer}: Hierarchical Vision Transformer using Shifted Windows},
year = {2021},
pages = {9992-10002},
doi = {10.1109/ICCV48922.2021.00986},
url = {https://doi.ieeecomputersociety.org/10.1109/ICCV48922.2021.00986},
publisher = {IEEE Computer Society}
}

@INPROCEEDINGS {li2021aha,
author = {Li, Jie and Ji, Rongrong and Chen, Peixian and Zhang, Baochang and Hong, Xiaopeng and Zhang, Ruixin and Li, Shaoxin and Li, Jilin and Huang, Feiyue and Wu, Yongjian},
booktitle = {IEEE/CVF International Conference on Computer Vision},
title = {Aha! Adaptive History-driven Attack for Decision-based Black-box Models},
year = {2021},
pages = {16148-16157},
doi = {10.1109/ICCV48922.2021.01586},
url = {https://doi.ieeecomputersociety.org/10.1109/ICCV48922.2021.01586},
publisher = {IEEE Computer Society},
}

@inproceedings{wang2022triangle,
author = {Wang, Xiaosen and Zhang, Zeliang and Tong, Kangheng and Gong, Dihong and He, Kun and Li, Zhifeng and Liu, Wei},
title = {Triangle Attack: A Query-Efficient Decision-Based Adversarial Attack},
year = {2022},
isbn = {978-3-031-20064-9},
publisher = {Springer-Verlag},
//address = {Berlin, Heidelberg},
url = {https://doi.org/10.1007/978-3-031-20065-6_10},
doi = {10.1007/978-3-031-20065-6_10},
booktitle = {European Conference on Computer Vision},
pages = {156--174},
//numpages = {19},
//location = {Tel Aviv, Israel}
}

@inproceedings{ma2021finding,
 author = {Ma, Chen and Guo, Xiangyu and Chen, Li and Yong, Jun-Hai and Wang, Yisen},
 booktitle = {Advances in Neural Information Processing Systems},
 //editor = {M. Ranzato and A. Beygelzimer and Y. Dauphin and P.S. Liang and J. Wortman Vaughan},
 pages = {19288--19300},
 //publisher = {Curran Associates, Inc.},
 title = {Finding Optimal Tangent Points for Reducing Distortions of Hard-label Attacks},
 //url = {https://proceedings.neurips.cc/paper_files/paper/2021/file/a113c1ecd3cace2237256f4c712f61b5-Paper.pdf},
 volume = {34},
 year = {2021}
}

@article{nesterov2017random,
  title={Random Gradient-Free Minimization of Convex Functions},
  author={Nesterov, Yurii and Spokoiny, Vladimir},
  journal={Foundations of Computational Mathematics},
  volume={17},
  number={2},
  pages={527--566},
  year={2017},
  doi={10.1007/s10208-015-9296-2},
  url={https://doi.org/10.1007/s10208-015-9296-2},
  issn={1615-3383}
}

@inproceedings{szegedy2017inceptionv4,
author = {Szegedy, Christian and Ioffe, Sergey and Vanhoucke, Vincent and Alemi, Alexander A.},
title = {{Inception-v4}, {Inception-ResNet} and the Impact of Residual Connections on Learning},
year = {2017},
publisher = {AAAI Press},
booktitle = {AAAI Conference on Artificial Intelligence},
pages = {4278--4284},
//numpages = {7},
//location = {San Francisco, California, USA},
series = {AAAI'17}
}

@inproceedings{chen2020rays,
author = {Chen, Jinghui and Gu, Quanquan},
title = {{RayS}: A Ray Searching Method for Hard-label Adversarial Attack},
year = {2020},
isbn = {9781450379984},
publisher = {Association for Computing Machinery},
//address = {New York, NY, USA},
//url = {https://doi.org/10.1145/3394486.3403225},
doi = {10.1145/3394486.3403225},
booktitle = {ACM SIGKDD International Conference on Knowledge Discovery and Data Mining},
pages = {1739–1747},
numpages = {9},
//location = {Virtual Event, CA, USA},
series = {KDD '20}
}

@inproceedings{chen2019hopskipjumpattack,
  author={Chen, Jianbo and Jordan, Michael I. and Wainwright, Martin J.},
  booktitle={IEEE Symposium on Security and Privacy}, 
  title={Hop{S}kip{J}ump{A}ttack: A Query-Efficient Decision-Based Attack}, 
  year={2020},
  pages={1277-1294},
  doi={10.1109/SP40000.2020.00045}
  }

@inproceedings{croce2021robustbench,
 author = {Croce, Francesco and Andriushchenko, Maksym and Sehwag, Vikash and Debenedetti, Edoardo and Flammarion, Nicolas and Chiang, Mung and Mittal, Prateek and Hein, Matthias},
 booktitle = {Proceedings of the Neural Information Processing Systems Track on Datasets and Benchmarks},
 //editor = {J. Vanschoren and S. Yeung},
 title = {{RobustBench}: a standardized adversarial robustness benchmark},
 url = {https://datasets-benchmarks-proceedings.neurips.cc/paper_files/paper/2021/file/a3c65c2974270fd093ee8a9bf8ae7d0b-Paper-round2.pdf},
 volume = {1},
 year = {2021}
}

@inproceedings{maho2021surfree,
author = {Maho, Thibault and Furon, Teddy and Le Merrer, Erwan},
booktitle = {IEEE/CVF Conference on Computer Vision and Pattern Recognition},
title = {{SurFree}: a fast surrogate-free black-box attack},
year = {2021},
pages = {10425-10434},
doi = {10.1109/CVPR46437.2021.01029},
url = {https://doi.ieeecomputersociety.org/10.1109/CVPR46437.2021.01029},
publisher = {IEEE Computer Society}
}

@inproceedings{dong2019efficient,
  author={Dong, Yinpeng and Su, Hang and Wu, Baoyuan and Li, Zhifeng and Liu, Wei and Zhang, Tong and Zhu, Jun},
  booktitle={IEEE/CVF Conference on Computer Vision and Pattern Recognition}, 
  title={Efficient Decision-Based Black-Box Adversarial Attacks on Face Recognition}, 
  year={2019},
  pages={7706-7714},
  doi={10.1109/CVPR.2019.00790}}

@inproceedings{cheng2019queryefficient,
title={Query-Efficient Hard-label Black-box Attack: An Optimization-based Approach},
author={Minhao Cheng and Thong Le and Pin-Yu Chen and Huan Zhang and JinFeng Yi and Cho-Jui Hsieh},
booktitle={International Conference on Learning Representations},
year={2019},
url={https://openreview.net/forum?id=rJlk6iRqKX},
}

@INPROCEEDINGS {li2020qeba,
author = {Li, Huichen and Xu, Xiaojun and Zhang, Xiaolu and Yang, Shuang and Li, Bo},
booktitle = {IEEE/CVF Conference on Computer Vision and Pattern Recognition},
title = {{QEBA}: Query-Efficient Boundary-Based Blackbox Attack},
year = {2020},
pages = {1218-1227},
doi = {10.1109/CVPR42600.2020.00130},
//address = {Los Alamitos, CA, USA},
//month =Jun,
url = {https://doi.ieeecomputersociety.org/10.1109/CVPR42600.2020.00130},
publisher = {IEEE Computer Society}
}

@misc{xu2023mimir,
	title={{MIMIR}: Masked Image Modeling for Mutual Information-based Adversarial Robustness},
	author={Xu, Xiaoyun and Yu, Shujian and Liu, Zhuoran and Picek, Stjepan},
	year={2025},
	eprint={2312.04960},
	archivePrefix={arXiv},
	primaryClass={cs.CV},
    url={https://arxiv.org/abs/2312.04960}, 
}
\end{document}